\documentclass[sigplan, nonacm]{acmart}

\usepackage{booktabs}
\usepackage{graphicx}
\usepackage{amsmath, amsfonts}
\usepackage{algorithm}
\usepackage{algpseudocode}
\usepackage{subcaption}
\usepackage{mathtools}
\usepackage{xcolor}
\usepackage{microtype}

\usepackage[percent]{overpic} %
\usepackage{hyperref}         %
\usepackage{xcolor}
\usepackage{pifont}

\title{\textsc{Tangram}: Unlocking Non-Uniform KV Cache Compression for Efficient Multi-turn LLM Serving}

\author{Hyungmin Kim}
\authornote{Equal contribution}
\affiliation{%
  \institution{Hanyang University}
  \city{Seoul}
  \country{Republic of Korea}
}
\email{kong4274@hanyang.ac.kr}

\author{Minsoo Kim}
\authornotemark[1]
\authornote{Currently at Apple}
\affiliation{%
  \institution{Hanyang University}
  \city{Seoul}
  \country{Republic of Korea}
}
\email{minsoo2333@hanyang.ac.kr}

\author{Hongseok Kim}
\affiliation{%
  \institution{Rebellions}
  \country{Republic of Korea}
}
\email{hongseok@rebellions.ai}

\author{Jungwook Choi}
\authornote{Corresponding author}
\affiliation{%
  \institution{Hanyang University}
  \city{Seoul}
  \country{Republic of Korea}
}
\email{choij@hanyang.ac.kr}
\begin{document}

\begin{abstract}

Multi-turn LLM serving accumulates dialogue history whose Key-Value (KV) cache grows with every turn and every user, quickly exceeding the model weights themselves and making memory---not compute---the binding constraint on throughput. Non-uniform KV compression, which allocates heterogeneous budgets across attention heads, preserves accuracy far better than uniform schemes, yet remains impractical: modern serving stacks assume identical KV lengths across heads, so heterogeneity traps freed memory as page fragmentation, spends up to 25\% of prefill time reclaiming scattered pages, and skews GPU workloads that inflate decode latency by up to 1.7$\times$ or burn 15--20\% of each decode step on re-planning. We observe that this heterogeneity need not be discovered at runtime: head-wise retention follows a two-level structural regularity---an input-invariant head ranking with narrowly bounded per-head ratios---that can be calibrated offline from as few as 50 samples. Building on this insight, we present \textsc{Tangram}, a serving framework that statically resolves what prior systems handle dynamically: \emph{Budget Reservation} fixes each head's post-compression footprint at scheduling time, eliminating page reclamation; \emph{Ragged Paging} clusters similar-budget heads into independent page tables, turning fragmentation into reclaimable memory; and \emph{Ahead-of-Time Load Balancing} precomputes balanced GPU partitions with zero runtime planning. Implemented on vLLM, \textsc{Tangram} serves as a drop-in substrate for existing non-uniform compression methods, matching their accuracy while improving end-to-end throughput by up to 2.6$\times$ over the full-KV baseline. Our implementation is publicly available at \url{https://github.com/aiha-lab/TANGRAM}.

\end{abstract}

\maketitle

\section{Introduction}

\begin{figure}[t]
    \centering
    \includegraphics[width=\linewidth]{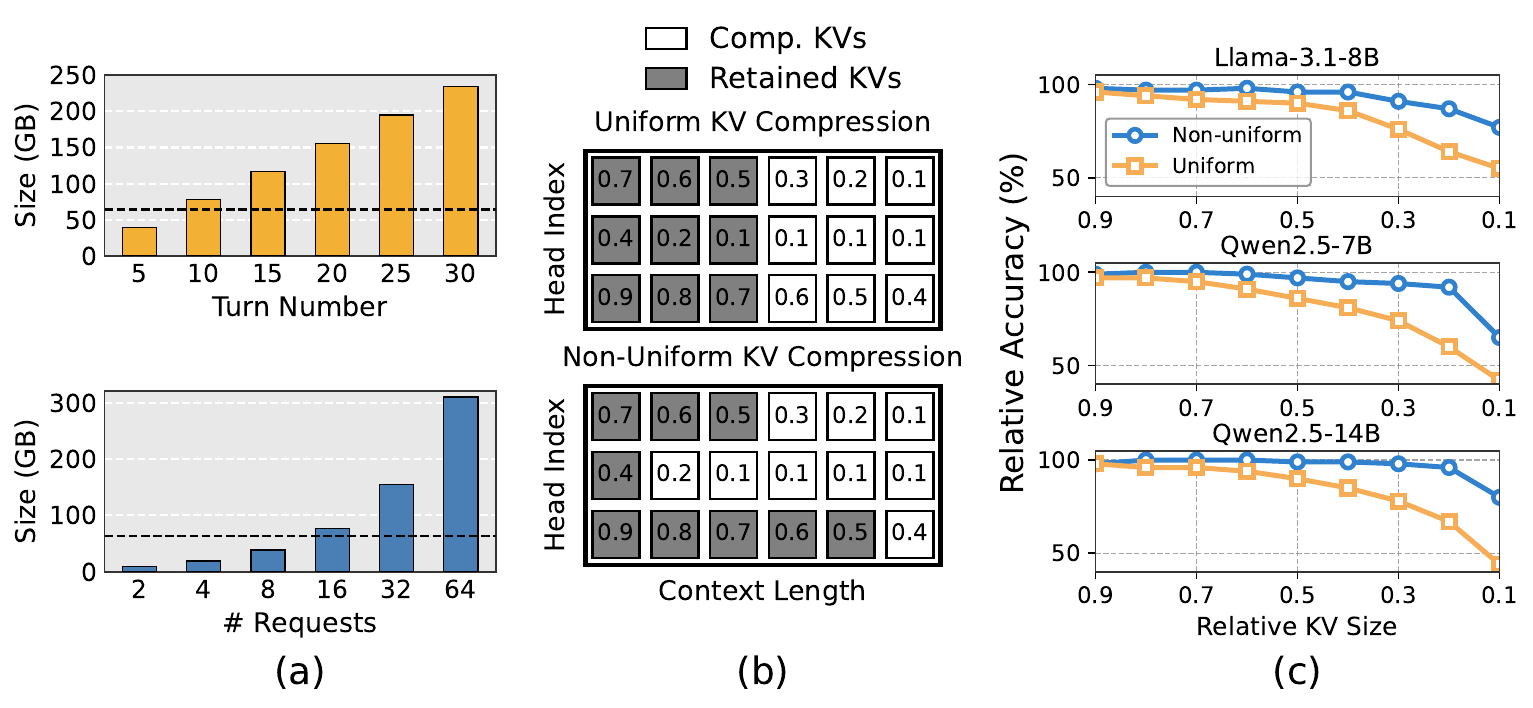}
    \caption{(a) KV cache size growth for Qwen2.5-32B with the number of conversation turns (top, \# requests = 16) or with the number of concurrent requests (bottom, \# turns = 10). The dashed line indicates the model weight size. (b) Comparison of uniform (top) and non-uniform (below) KV compression strategies at a 50\% target global retention ratio, where the numbers in each box denote the importance score of each KV entry. (c) Comparative accuracy on long-term conversation QA benchmarks~\cite{lee2025realtalk} using KVzip~\cite{kim2025kvzip} with Uniform and non-uniform KV compression.}
    \label{fig:kvcache_size}
\end{figure}

Multi-turn interaction has become the dominant way users engage with Large Language Models (LLMs): AI assistants now accumulate dialogue history across sessions to deliver consistent, personalized responses~\cite{li2025singleturnsurveymultiturninteractions,anthropic2024claudememory,achiam2023gpt,openai2024memory}. To avoid re-computing this ever-growing history $\mathcal{H}_t$ at every turn~\cite{lee2025realtalk,maharana-etal-2024-evaluating,wu2025longmemeval,gorle2025quantifying}, serving systems persist attention states in the Key-Value (KV) cache~\cite{pope2023efficiently}---but the cache grows linearly with every turn and every concurrent user. For Qwen2.5-32B, with merely 16 concurrent requests, the accumulated KV cache surpasses the size of the model weights themselves within ten conversation turns, and continues to grow unboundedly thereafter (Figure~\ref{fig:kvcache_size}(a)). In multi-turn serving, memory capacity---not compute---is the binding constraint on batch size and therefore on throughput.

KV cache compression is the standard remedy, but \emph{how} the token budget is distributed across attention heads determines whether accuracy survives. \emph{Uniform} compression~\cite{jiang2023mistral7b,xiao2024efficient,oren-etal-2024-transformers,li2024snapkv,zhang2023ho,kim-etal-2024-infinipot} forces every head to retain the same number of tokens, ignoring a fundamental property of attention: critical long-range information is concentrated in a small subset of \emph{retrieval heads}~\cite{fu2025not}, while other heads attend only locally. Truncating all heads equally starves precisely the heads that matter, and accuracy degrades sharply in multi-turn settings (Figure~\ref{fig:kvcache_size}(c)). \emph{Non-uniform} compression~\cite{feng2024ada,xiao2025duoattention,fu2025not,tang2025razorattention,kim2025kvzip} instead allocates heterogeneous per-head budgets that mirror this skew, preserving near-original accuracy even under aggressive compression. Thus, non-uniform compression is the right tool for memory-efficient multi-turn serving.

Systemically, however, non-uniform compression is impractical today. The software stack of state-of-the-art serving systems---PagedAttention~\cite{kwon2023efficient}, continuous batching with chunked prefill~\cite{agrawal2023sarathi, yu2022orca}, and optimized attention kernels~\cite{dao2023flashattention,dao_flashdecoding, ye2025flashinfer}---is architected end-to-end around a single implicit assumption: \emph{all attention heads hold KV caches of identical length}. Non-uniform compression violates this assumption at every layer of the stack, resulting in severe overhead. First, the monolithic page of PagedAttention~\cite{kwon2023efficient} spans all heads at a single uniform length, so the heterogeneous per-head budgets of non-uniform compression cannot be realized on it in the first place; the memory that compression would free instead stays trapped as unrecoverable \emph{page fragmentation} (\S~\ref{sec:page_fragmentation}). Second, because each head's post-compression footprint is unknown until the forward pass computes importance scores, the scheduler must over-allocate and then track, reclaim, and remap scattered pages in flight; this control-plane churn consumes up to 25\% of prefill execution time (\S~\ref{sec:scheduling_uncertainty}, Figure~\ref{fig:latency_breakdown}). Third, heterogeneous per-head KV lengths skew the workload across GPU SMs: static kernel partitioning~\cite{dao_flashdecoding} suffers stragglers that inflate decode attention latency by up to 1.7$\times$, while dynamic per-layer re-planning~\cite{ye2025flashinfer} burns 15--20\% of every decode step on the CPU (\S~\ref{sec:workload_imbalance}). In short, the theoretical memory savings of non-uniform compression evaporate inside the serving system---and can even regress end-to-end throughput.

Rather than treating these runtime overheads as inevitable, our profiling reveals a crucial insight: head-wise KV retention exhibits a \emph{two-level structural regularity} that is intrinsic to the model rather than driven by the input (Figure~\ref{fig:kvsize}). Specifically, the \emph{ranking} of attention heads by retention demand is essentially input-invariant, and each head's \emph{absolute} retention ratio varies only within a narrow, estimable band. This regularity fundamentally changes the problem landscape. KV cache heterogeneity need not be discovered dynamically; it can be \emph{calibrated once, offline}, and treated as a static blueprint. Every runtime burden imposed by non-uniformity---page bookkeeping, reclamation, and kernel planning---becomes a deterministic decision resolvable before execution. Notably, prior heterogeneous memory systems either manage heterogeneity at coarser layer granularity~\cite{zhang2025jenga} or
treat each head independently~\cite{zhang2025diffkv}; neither
exploits this cross-head structural regularity.

Building on this insight, we present \textsc{Tangram}, a serving framework that turns non-uniform KV compression from an algorithmic promise into realized system performance. \textsc{Tangram} resolves each of the three bottlenecks with a dedicated, statically planned mechanism. \emph{Budget Reservation} (\S\ref{sec:fixed_budget}) fixes each head's budget to its offline-calibrated value, so the scheduler reserves exactly the post-compression pages at scheduling time---eliminating over-allocation and the entire compress-and-reclaim path. \emph{Ragged Paging} (\S\ref{sec:grouped_paging}) replaces the monolithic page with finer-grained, per-group page tables, clustering heads of similar budget so that freed capacity is physically reclaimable rather than trapped; a vectorized block table keeps the added control-plane cost negligible. \emph{Ahead-of-Time (AOT) Load Balancing} (\S\ref{sec:load_balancing}) precomputes a Workload Split Map from the reserved budget profiles, delivering balanced SM utilization without runtime planning. A natural concern is that freezing budgets offline sacrifices the input-adaptivity that makes non-uniform compression accurate; our evaluation shows the opposite: with a small calibrated safety margin, \textsc{Tangram} matches---and occasionally exceeds---the accuracy of the original dynamic implementations across five models and three state-of-the-art compression methods (\S\ref{sec:ltc_acc_eval}).

In summary, this paper makes the following contributions:
\begin{itemize}
  \item \textbf{Characterization.} We establish a two-level structural regularity in head-wise KV retention---input-invariant head rankings with narrowly bounded ratios---enabling offline calibration of non-uniform compression profiles from as few as 50 samples (\S\ref{key_observation}).

  \item \textbf{Co-designed system.} Exploiting this regularity, \emph{Budget Reservation}, \emph{Ragged Paging}, and \emph{AOT Load Balancing} statically resolve what prior systems handle at runtime: eliminating page reclamation (up to 25\% of prefill time), reclaiming 12--25\% more memory, and removing per-layer kernel planning (\S\ref{sec:method}).

  \item \textbf{Generality and results.} Built on vLLM, \textsc{Tangram} is a drop-in substrate for existing non-uniform methods~\cite{feng2024ada, devoto2025expected, kim2026fast}, preserving their accuracy while improving end-to-end throughput by up to \textbf{2.6$\times$} over the full-KV baseline (\S\ref{sec:eval-perf}). Our implementation is publicly available.
\end{itemize}

\section{Background}

\label{sec:background}

\subsection{Non-uniform KV Cache Compression}%
\label{ssec:non-uniform_kv}

\subsubsection{KV Cache Bottleneck in Multi-Turn LLMs}

Multi-turn interactions have emerged as the dominant LLM workload, where a model must engage with users over extended periods while maintaining contextual coherence. We formalize each exchange as an interaction unit $(u_i, a_i)$, consisting of a user utterance $u_i$ and the corresponding model response $a_i$ in a \textit{token} sequence. The system then maintains a cumulative dialogue history $\mathcal{H}_t = \{(u_i, a_i)\}_{i=1}^{t-1}$ for each user request, which serves as the essential context for generating the response at turn $t$~\cite{wu2025longmemeval,hu2026evaluating}.

Serving systems maintain this history with the Key-Value (KV) cache, which incrementally stores attention states to avoid redundant re-computation of $\mathcal{H}_t$ \cite{pope2023efficiently}. For an $L$-layer, $H$-head Transformer, this requires storing Key and Value tensors for every token across all layers and heads, causing the cache size to scale with the length of the accumulated dialogue~\cite{ghadia2025dialogue,kim2025epicacheepisodickvcache}. Throughout this paper, $H$ denotes the number of \emph{KV heads}: under grouped-query attention (GQA), multiple query heads share one KV head, and KV cache compression operates at KV-head granularity (e.g., $H{=}8$ for Llama-3.1-8B). As the number of concurrent user requests (i.e., batch size) grows and $\mathcal{H}_t$ accumulates across turns, this scaling pressure compounds rapidly. As illustrated in Figure~\ref{fig:kvcache_size}(a), the KV cache footprint often surpasses the model size even with a few concurrent requests. Consequently, memory capacity---rather than compute---becomes the primary constraint on system throughput, necessitating efficient compression strategies.

\subsubsection{KV Cache Compression}
KV compression reduces the cache footprint by retaining only the most \textit{critical} tokens per head---those that receive high cumulative attention weights and thus contribute most to the attention output. Formally, for a context of $N$ tokens, the \textit{importance score} $s_{\ell,h} \in \mathbb{R}^N$ aggregates the attention weights each token receives at head $h$ in layer $\ell$. Given a \textit{target global retention ratio} $\rho \in (0,1]$ (the \emph{kept-cache fraction}), compression selects a set of retained token indices $I_{\ell,h}$ and constructs the compressed KV cache accordingly:
\begin{equation}
    \widetilde{K}_{\ell,h} = K_{\ell,h}[I_{\ell,h}, :], \qquad \widetilde{V}_{\ell,h} = V_{\ell,h}[I_{\ell,h}, :].
\end{equation}
A fundamental property of the attention mechanism, however, is that heads exhibit \textit{diverse concentration patterns}: some heads sharply concentrate their attention weights on a small subset of tokens, while others distribute them broadly across the context~\cite{wu2025retrieval,xiao2025duoattention}, causing the number of critical tokens to vary substantially across heads. How $I_{\ell,h}$ is computed under $\rho$ leads to two fundamentally different compression strategies.

\textit{Uniform KV compression}~\cite{zhang2023ho,oren-etal-2024-transformers,li2024snapkv} ignores head-wise diversity by applying a fixed per-head token budget $\lceil \rho N \rceil$ identically to every head, as shown in the upper panel of Figure~\ref{fig:kvcache_size}(b). Compression then selects the top-$\lceil \rho N \rceil$ tokens independently for each head:
\begin{equation}
    I_{\ell,h} = \mathrm{Top}(\lceil \rho N \rceil,\; s_{\ell,h}).
\end{equation}
However, as shown in the lower panel of Figure~\ref{fig:kvcache_size}(b), heads naturally exhibit heterogeneous importance distributions across the context, so applying a uniform token budget indiscriminately truncates each head's context regardless of its actual distribution, discarding tokens that are critical for broadly-attending heads while over-retaining tokens for narrowly-attending ones.

\textit{Non-uniform KV compression}~\cite{feng2024ada,fu2025not,kim2025kvzip} removes the per-head budget constraint by enforcing $\rho$ at the layer level. It flattens importance scores across all $H$ heads ($s_{\ell}^{\mathrm{flat}} = \mathrm{concat}_h(s_{\ell,h}) \in \mathbb{R}^{HN}$) and applies a single global token budget $\lceil \rho H N \rceil$:
\begin{equation}
    I_{\ell} = \mathrm{Top}(\lceil \rho H N \rceil,\; s_{\ell}^{\mathrm{flat}}).
\end{equation}
This yields a highly irregular \textit{implicit per-head retention ratio}---some heads retain their full history while others are heavily pruned---naturally mirroring the heterogeneous concentration patterns of attention. However, since this per-head ratio is determined by the global top-$\lceil \rho HN \rceil$ selection over flattened scores, it varies with every input and is unknown before the forward pass. Crucially, Figure~\ref{fig:kvcache_size}(c) demonstrates that this head-wise budget heterogeneity preserves high conversational accuracy even under aggressive KV size compression, establishing non-uniform KV compression as essential for memory-efficient multi-turn LLM serving.

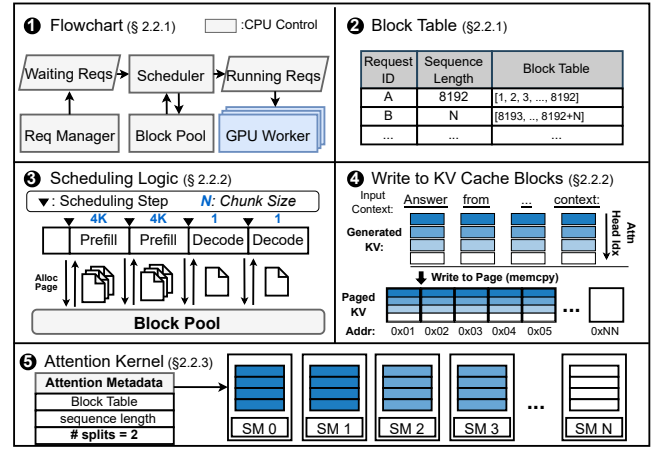
\begin{figure}[t]
    \centering
    \begin{overpic}[width=\linewidth,]{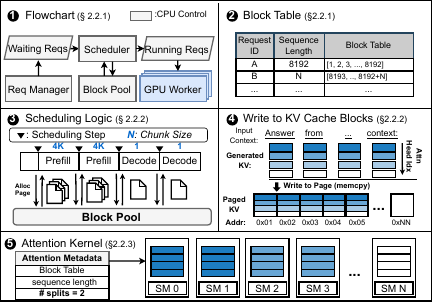}

        \put(20, 65) {\hyperref[ssec:continuous_batching]{\color{red}\makebox(7,3){}}}
        \put(70, 65) {\hyperref[ssec:continuous_batching]{\color{red}\makebox(7,3){}}}

        \put(30, 42) {\hyperref[paged_attention]{\color{blue}\makebox(7,3){}}}
        \put(77, 42) {\hyperref[paged_attention]{\color{blue}\makebox(7,3){}}}

        \put(25,13) {\hyperref[attn_kernel_optimization]{\color{green}\makebox(7,3){}}}

    \end{overpic}
    \caption{Main components of vLLM.}
    \label{fig:explain_vllm}
\end{figure}

\subsection{LLM Serving System}
\label{ssec:serving_system}

State-of-the-art serving frameworks such as vLLM~\cite{kwon2023efficient} and SGLang~\cite{zheng2023efficiently} rely on a tightly integrated execution pipeline to manage memory and compute. As illustrated in Figure \ref{fig:explain_vllm}, this pipeline is driven by four core components: a scheduler, a block table, KV cache blocks (pages), and the attention kernel, which together enable efficient batching and memory management~\cite{agrawal2023sarathi,patel2024splitwise,pope2023efficiently,yu2022orca,zhong2024distserve,lee2024infinigen}. Crucially, this entire system structure is built under the implicit assumption that KV cache lengths remain \textit{uniform} across all attention heads.

\subsubsection{Continuous Batching.}
\label{ssec:continuous_batching}

The scheduler manages the lifecycle of incoming and active user requests through iteration-level continuous batching~\cite{yu2022orca}. At each scheduling step, it inspects the current status of the Block Pool of requests to make decisions on admission and execution (\ding{182}). Once a request is considered runnable, the scheduler allocates physical pages to accommodate its KV cache via the Block Table, which maps physical page addresses to specific Request IDs (\ding{183}). Modern serving systems pair continuous batching with \emph{chunked prefill}~\cite{agrawal2023sarathi}, which splits a long context prefill request into fixed-size token chunks processed over successive scheduling steps and interleaved with other requests' decode steps (\ding{184}). By preventing any single long-context prefill request from monopolizing an iteration, it bounds the time-to-first-token (TTFT) for concurrent requests, making it indispensable for high-throughput serving. 
For every iteration, the scheduler allocates pages, computes their
physical addresses, and adjusts overall KV usage on the host CPU as
part of the control plane---all under the implicit assumption of a
static, uniform per-token memory cost.

\subsubsection{PagedAttention.}
\label{paged_attention}

In the LLM Forward Stage (\ding{184}--\ding{185} in Figure~\ref{fig:explain_vllm}), the GPU worker writes the generated KV entries into KV cache blocks. Between consecutive forward passes, a scheduling step requests the pages required for the next pass from the Block Pool and precomputes their physical addresses, which the forward pass then uses to store each KV entry non-contiguously in fixed-size blocks, eliminating \emph{external} memory fragmentation across requests. A key design constraint of PagedAttention~\cite{kwon2023efficient} is its unified page structure: a single physical block spans all layers and all attention heads simultaneously, holding $L \times H \times 2 \times P \times d$ elements for $P$ consecutive tokens (the factor 2 for Key and Value, $d$ the per-head dimension), making granular head-wise compression impossible. As we show in \S\ref{sec:page_fragmentation}, this very structure creates a new, \emph{internal} form of fragmentation once per-head KV lengths diverge.

\subsubsection{Attention Kernel Optimization.}
\label{attn_kernel_optimization}

FlashAttention-2~\cite{dao2023flashattention} reduces redundant HBM--SRAM traffic via tiled, fused attention computation along the query dimension. For long-context decoding, FlashDecoding~\cite{dao_flashdecoding} and FlashInfer~\cite{ye2025flashinfer} further introduce KV-dimension parallelism (\ding{186} in Figure~\ref{fig:explain_vllm}), where the \emph{number of splits} determines how each attention head is partitioned and distributed across SMs during decode attention. While FlashDecoding~\cite{dao_flashdecoding} relies on static heuristics for partitioning, FlashInfer~\cite{ye2025flashinfer} employs a runtime planning phase to identify optimal workload strategies. This planning cost can be amortized through \emph{plan reuse}: since all layers typically share identical KV structures, the system computes a single plan and reuses it across all $L$ layers to reduce planning overhead.

\begin{figure}[t]
    \centering
    \includegraphics[width=\linewidth]{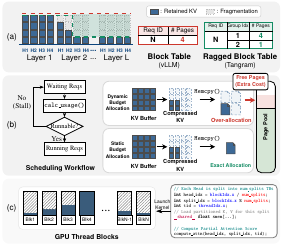}
    \caption{Challenges posed by non-uniform KV compression. (a) \textit{Monolithic Page Structure}: unified pages span all heads, causing page fragmentation (red dashed: pages allocated per request; teal dashed: per-group allocation under Ragged Paging). (b) \textit{Page Management Overhead}: reclaiming scattered pages at runtime incurs severe control-plane cost. (c) \textit{Workload Imbalance}: uniform KV splits across thread blocks cause stragglers under different per-head KV lengths.}
    \label{fig:challenges}
\end{figure}

\section{Motivation}
\label{sec:challenges}

While non-uniform KV compression preserves multi-turn accuracy, deploying it on production systems such as vLLM~\cite{kwon2023efficient} reveals severe inefficiencies: every pillar of the serving stack described in \S\ref{ssec:serving_system} assumes uniform per-head KV lengths. We analyze the three
resulting limitations in turn, each motivating one technique in \S\ref{sec:method}.

\subsection{Limitations of Existing Systems}
\label{limitations_on_existing_system}

\subsubsection{Monolithic Page Structure.}
\label{sec:page_fragmentation}

The first limitation arises directly from PagedAttention's monolithic page structure (\S\ref{paged_attention}). Figure~\ref{fig:challenges}(a) illustrates the problem: under non-uniform compression, each head retains a different number of KV entries (filled blocks), yet the block table records only a single page count per request, so every head in every layer is allocated up to the longest-retaining head (red dashed line). The slots between a head's actual retention and this allocation ceiling (gray blocks) hold entries that are already evicted but can never be returned to the page pool, because a page is shared by all heads and freed only when every head releases it. We term this \emph{page fragmentation}. \textit{Ragged Paging} (\S\ref{sec:grouped_paging}) removes this coupling by managing heads with similar retention in independent page tables, whose per-group allocation (teal dashed line) tracks each group's own maximum and reclaims the fragmented capacity.

\subsubsection{Page Management Overhead}
\label{sec:scheduling_uncertainty}

The second limitation stems from the interaction between non-uniform compression and the scheduler's control plane (\S\ref{ssec:continuous_batching}). As illustrated in Figure~\ref{fig:challenges}(b), at each scheduling step the scheduler inspects the block pool, promotes waiting requests to the running set, and allocates the pages each request needs for the current forward step. Under non-uniform compression, however, how much each attention head's KV cache will be compressed---and thus how many pages it needs---is unknown until the forward pass computes token importance scores at runtime. The scheduler must therefore over-allocate and then run a costly ``compress-and-reclaim'' process: identifying the scattered pages freed by compression, returning them to the block pool, and updating page tables while the request is in flight. This overhead scales linearly with the number of reclaimed pages, consuming up to 25\% of total prefill execution time (Figure~\ref{fig:latency_breakdown}) and directly limiting throughput.

\subsubsection{Workload Imbalance}
\label{sec:workload_imbalance}

The third limitation occurs at the GPU kernel level during decode attention. GPU architectures achieve peak efficiency under the SIMT paradigm only when parallel threads process uniform workloads. Non-uniform KV compression breaks this uniformity in two distinct ways.

\begin{figure}[t]
    \centering
    \includegraphics[width=\linewidth]{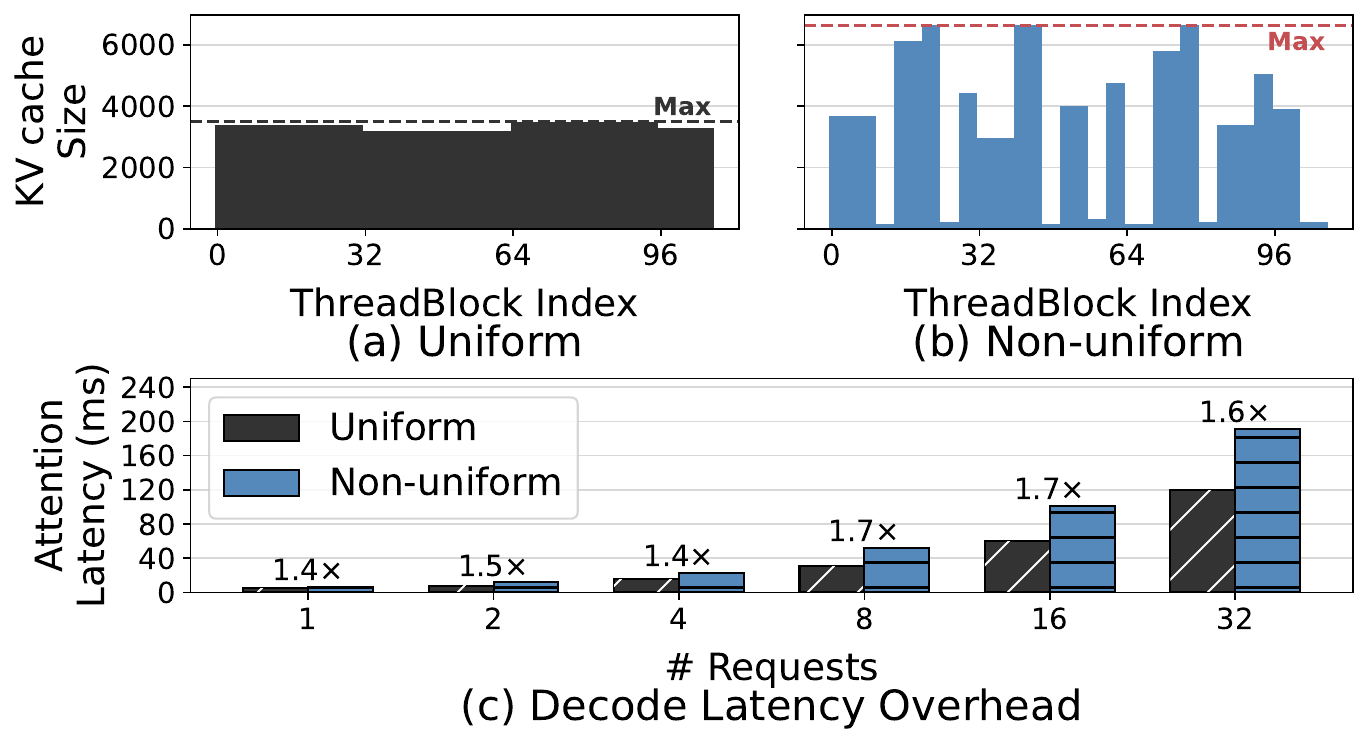}
    \caption{Workload imbalance on decode attention. (a) Uniform KV compression, (b) Non-uniform KV compression, (c) attention latency across different configurations of requests on Qwen3-4B. The dashed line indicates the maximum workload among all thread blocks.}
    \label{fig:workload_imbalance}
\end{figure}

\paragraph{Straggler Effect from Static Partitioning.}
As shown in Figure~\ref{fig:challenges}(c), FlashDecoding~\cite{dao_flashdecoding} parallelizes attention by dispatching fixed-size KV chunks to GPU SMs based on a \texttt{num\_splits} parameter applied uniformly across all heads---a heuristic valid only when every head has the same context length. Under non-uniform compression, per-head KV lengths can differ severely across heads. As illustrated in Figure~\ref{fig:workload_imbalance}(a--b), this heterogeneity produces highly skewed per-thread-block workloads: blocks mapped to long-context heads become \emph{heavy} while short-context blocks finish early and idle. The overall decoding step is gated by the few SMs executing the heaviest blocks, increasing decode attention latency by up to $1.7\times$ compared to the uniform baseline under the same total KV cache size (Figure~\ref{fig:workload_imbalance}(c)).

\begin{table}[t]
    \centering
    \footnotesize
    \setlength{\tabcolsep}{5pt}
    \begin{tabular}{lccc}
        \toprule
        Model & \# Layers & Load Balancing Time (ms) & Proportion \\
        \midrule
        Qwen2.5-7B    & 28 &  5.64 & 20.20\% \\
        Qwen3-32B     & 64 & 13.27 & 15.48\% \\
        Llama-3.1-8B  & 32 &  6.48 & 17.33\% \\
        Llama-3.1-70B & 80 & 17.13 & 16.43\% \\
        \bottomrule
    \end{tabular}
    \caption{Load balancing overhead. Proportion denotes the fraction of the total decode inference step time spent on the load balancing phase.}
    \label{tab:load_balancing_overhead}
\end{table}

\paragraph{Prohibitive Cost of Dynamic Rebalancing.}
FlashInfer~\cite{ye2025flashinfer} addresses workload imbalance through a runtime planning phase before each decoding step. Under uniform settings, \emph{plan reuse} amortizes this cost: a single plan is computed once and reused across all $L$ layers. Non-uniform compression invalidates this optimization---retained KV lengths differ independently across layers, forcing the planner to recompute a unique partition \emph{for every layer} at every decoding step. As shown in Table~\ref{tab:load_balancing_overhead}, this per-layer planning overhead consumes 15--20\% of the total decode iteration time, negating the GPU utilization gains that dynamic balancing is meant to provide.

\subsection{Key Observation and System Design}
\label{key_observation}

Prior work has established that attention heads in Transformer-based models exhibit model-intrinsic, specialized roles that persist regardless of the input~\cite{voita-etal-2019-analyzing, NEURIPS2019_2c601ad9, wu2025retrieval, xiao2025duoattention}. Our empirical analysis reveals that this intrinsic specialization is directly reflected in the KV retention profile, exhibiting a \emph{two-level structural regularity}: (1) the \emph{ranking} of attention heads by retention demand is essentially input-invariant, and (2) each head's \emph{absolute} retention ratio varies only within a narrow, estimable range. This implies that KV cache heterogeneity is not fundamentally unpredictable, but a deterministic structure that can be calibrated offline as a static blueprint for scheduling, memory management, and kernel execution.
\begin{figure}[t]
    \centering
    \includegraphics[width=\linewidth]{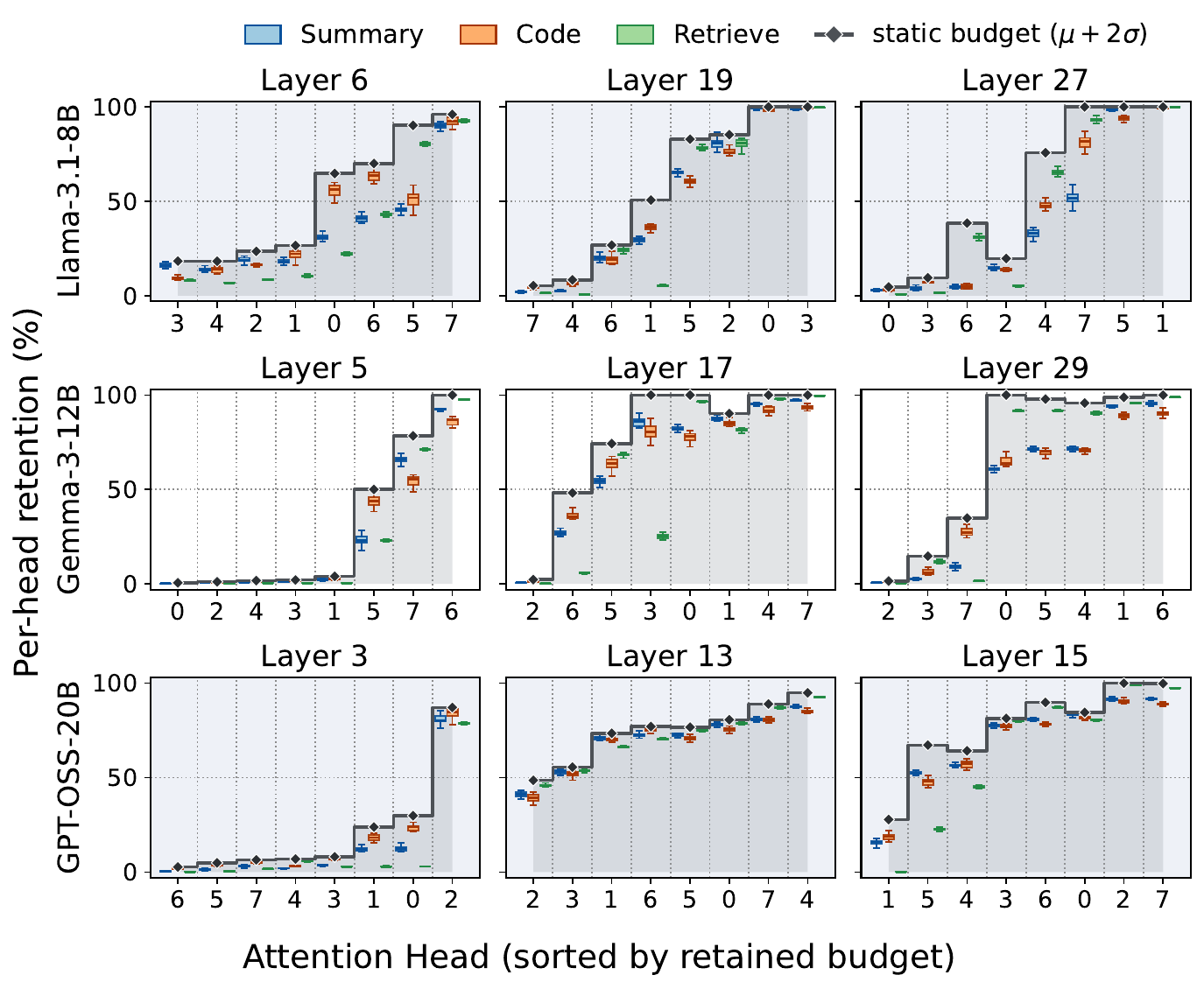}
    \caption{Per-head KV retention rates (\%) under non-uniform compression using KVzip~\cite{kim2025kvzip} at a 50\% target global retention ratio, across three model families (Llama-3.1-8B, Gemma-3-12B, GPT-OSS-20B) and three SCBench~\cite{li2024scbench} tasks---summarization (Summary), code understanding (Code; RepoQA), and fact retrieval (Retrieve; QA-Eng)---shown for three selected layers per model; for the hybrid models (Gemma-3-12B, GPT-OSS-20B), these are full-attention layers. Heads on the x-axis are sorted by their retained budget. Each box aggregates 50 input samples. Dashed lines mark the per-head budget that \textsc{Tangram} reserves with the safety coefficient $\alpha=2$ (\S\ref{sec:calibration}).}
    \label{fig:kvsize}
\end{figure}

As shown in Figure~\ref{fig:kvsize}, per-head retention rates are highly heterogeneous within a layer, yet each head exhibits a largely stable retention level across inputs---the narrow box widths for each head indicate low variance across the 50 samples. While the absolute retention values may shift moderately across tasks---spanning representative long-context subtasks \citep{li2024scbench} such as summarization (Summary), code QA (Code), and fact retrieval (Retrieve)---each head's relative retention level within a layer is consistently preserved, and this pattern holds across diverse models. Note that the heads in Figure~\ref{fig:kvsize} are ordered by their calibrated budget: the per-task markers of all three workloads rise largely monotonically along this single ordering, rather than reshuffling it, directly visualizing that diverse workloads induce the same head ranking. The low per-head variance further implies that this profile can be estimated reliably offline from only a handful of calibration samples---50 in all our experiments (\S\ref{sec:eval-setup})---and generalizes beyond them. While Figure~\ref{fig:kvsize} uses KVzip as the scoring method, this regularity is not an artifact of a particular method: repeating the same analysis for various non-uniform compression methods reproduces the same two-level structure for every method (Appendix~\ref{sec:appendix_methods}).

This stable, model-intrinsic retention structure converts all three sources of overhead into statically resolvable properties: each head's retention ratio can be fixed offline rather than recomputed per input, and heads with similar stable retention levels can be grouped once per model. Together, these enable three deterministic, pre-scheduled design decisions:
\begin{enumerate}
    \item \textbf{Budget Reservation} (\S\ref{sec:fixed_budget}): Fix each head's retention ratio offline, letting the scheduler reserve exactly the required pages before execution and eliminating the runtime compress-and-reclaim step.
    \item \textbf{Ragged Paging} (\S\ref{sec:grouped_paging}): Cluster heads with similar retention levels offline into independent page tables, enabling true physical memory reclamation and breaking monolithic page fragmentation.
    \item \textbf{Ahead-of-Time (AOT) Load Balancing} (\S\ref{sec:load_balancing}): Precompute optimal GPU workload partitions offline based on the fixed head group shapes, achieving balanced SM utilization with zero runtime planning overhead.
\end{enumerate}

\begin{figure*}[t]
    \centering
    \includegraphics[width=0.93\linewidth]{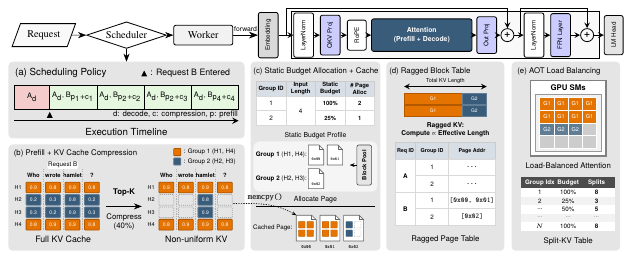}
    \caption{System overview of \textsc{Tangram}. \textbf{(a)} chunk-wise prefill with per-chunk compression at scheduling time; \textbf{(b)} a scoring function rates each KV entry to drive compression; \textbf{(c)} the static per-head budget fixes the post-compression memory footprint; \textbf{(d)} a ragged block table manages each head group at its own length; \textbf{(e)} an ahead-of-time workload partition balances decode attention across SMs.}
    \label{fig:system_overview}
\end{figure*}

\section{Methodology}
\label{sec:method}
We present \textbf{\textsc{Tangram}}, a holistic serving framework that reconciles the theoretical efficiency of non-uniform KV cache compression with the practical constraints of high-throughput serving. Throughout, compression remains fully fused with chunked prefill and continuous batching rather than reducing memory in isolation. Figure~\ref{fig:system_overview} illustrates how the three stages compose.

\begin{figure}[t]
    \centering
    \includegraphics[width=\linewidth]{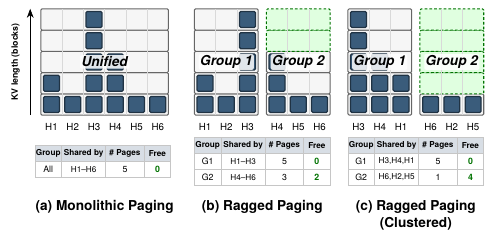}
    \caption{Freeing KV pages under non-uniform KV compression.
    A page is freed only when \emph{all} heads sharing it are evicted.
    \textbf{(a) Monolithic} shares one page across all heads, so a single long
    head (H3) blocks every page.
    \textbf{(b) Ragged} pages each head group ($H_p{=}3$) independently, freeing
    pages in groups without a long head.
    \textbf{(c) Clustered} first sorts heads by retention, packing short heads
    together to free far more pages.}
    \label{fig:paging}
\end{figure}

\subsection{Budget Reservation}
\label{sec:fixed_budget}
Our key observation (\S\ref{key_observation}) shows that each attention head's retention follows a model-intrinsic pattern, confined to a narrow, input-invariant range. \textsc{Tangram} leverages this regularity to fix every head's budget before execution, replacing dynamic, input-dependent compression with a statically planned memory footprint.

\paragraph{Fusing Compression into Prefill.}
When a request arrives, \textsc{Tangram} executes compression as an integral part of the prefill phase rather than as a separate post-processing pass. As shown in Figure~\ref{fig:system_overview}(a), each prefill chunk is compressed as soon as its KV cache is generated, fully fusing compression into chunked prefill and continuous batching---the backbone of modern serving (\S\ref{ssec:serving_system}). Within each chunk, the compression method's scoring function rates the importance of every KV entry (Figure~\ref{fig:system_overview}(b)). Rather than performing the global top-$\lceil \rho HN \rceil$ selection over flattened scores that yields a single index set $I_{\ell}$ (\S\ref{ssec:non-uniform_kv}), \textsc{Tangram} replaces the input-dependent implicit per-head ratio with a static \textit{per-head budget ratio} $B_{\ell,h} \in (0,1]$, decomposing compression into per-head top-$\lceil B_{\ell,h} N \rceil$ selections:
\begin{equation}
    I_{\ell,h} = \mathrm{Top}(\lceil B_{\ell,h} N \rceil,\; s_{\ell,h}).
    \label{eq:topk}
\end{equation}

\paragraph{Budget Calibration.}
\label{sec:calibration}
The per-head budget ratio $B_{\ell,h}$ is calibrated once, offline. Running non-uniform compression on a small set of sample contexts under a target global retention ratio $\rho$, we record the implicit per-head retention ratio each head receives from the global top-$\lceil \rho HN \rceil$ selection, summarized by its mean $\mu_{\ell,h}$ and standard deviation $\sigma_{\ell,h}$. Rather than recomputing this ratio for every input, we fix it with a controlled safety margin:
\begin{equation}
    B_{\ell,h} = \min\!\bigl(1,\;\mu_{\ell,h} + \alpha \cdot \sigma_{\ell,h}\bigr),
\end{equation}
where $\alpha$ is a safety-margin coefficient. By construction, the per-head budgets sum to approximately the global token budget: $\sum_{h} \lceil B_{\ell,h} N \rceil \approx \lceil \rho H N \rceil$, preserving the same overall retention level as the original non-uniform compression. As shown in Figure~\ref{fig:system_overview}(c), calibration thereby determines, for every head, both the number of pages it receives during prefill and the head group it is managed with (\S~\ref{sec:grouped_paging}). Since \textsc{Tangram} only fixes the per-head budget $B_{\ell,h}$ and leaves each method's scoring function untouched, it is directly compatible with the importance-scoring function of any KV compression method~\cite{li2024snapkv, park2026keydiff, devoto2025expected, kim2025kvzip, kim2026fast}---each method retains its own importance scoring, while \textsc{Tangram} calibrates the resulting per-head ratios independently into a static profile entirely offline, adding no cost to the serving path.

\paragraph{Precise Page Allocation.}
Since every per-head budget ratio $B_{\ell,h}$ is fixed, a request's post-compression footprint is fully determined before execution. The scheduler, therefore, reserves exactly the required number of pages when it admits the request: allocation occurs once, at scheduling time, and every page enters the cache already in its post-compression state. This eliminates the over-provisioning and scattered-page reclamation that underlie the \emph{Page Management Overhead} (\S\ref{sec:scheduling_uncertainty}), thereby removing its TTFT cost from the serving path entirely.

\paragraph{Robustness and Overflow Handling.} Figure~\ref{fig:kvsize} overlays the reserved budget $B_{\ell,h}$ on the observed retention distributions: it covers each head's per-input deviation---across all three workloads with distinct attention behaviors---while spending only a marginal amount of extra budget, indicating that calibration captures a model-intrinsic structure rather than the distribution of its pilot samples. The reserved budget is nonetheless a hard capacity bound: if an input demands more retention than $B_{\ell,h}$ for some head, the per-head top-$\lceil B_{\ell,h}N\rceil$ selection (Eq.~\ref{eq:topk}) simply retains the highest-scoring entries within capacity---equivalent to running the underlying method at a marginally lower ratio for that head---so overflow degrades gracefully into slightly stronger compression, never into a stall or a correctness failure. The end-to-end accuracy results in \S\ref{sec:ltc_acc_eval} (Figure~\ref{fig:acc}) already include any such residual truncation.

\subsection{Ragged Paging}

\label{sec:grouped_paging}

\textsc{Tangram} stores the KV cache as a \emph{ragged} structure: each head group is kept at its own retained length rather than padded to a uniform one---analogous to a ragged tensor such as TensorFlow's \texttt{RaggedTensor}, whose rows need not share a common length.

\subsubsection{Paging at Head-Group Granularity}
\label{sec:method:grouping}

To realize this structure, \textsc{Tangram} narrows the unit of paging from all attention heads jointly to the \emph{head group}: a set of $H_p$ attention heads whose KV cache is managed by a single page table, where $H_p$ (i.e., the number of heads sharing one page table) sets the granularity. Under this definition, conventional PagedAttention degenerates to a single group spanning all $L \times H$ heads, which is precisely why a monolithic page cannot release the memory freed by compression (Figure~\ref{fig:paging}(a)).

\begin{figure}[t]
    \centering
    \includegraphics[width=\linewidth]{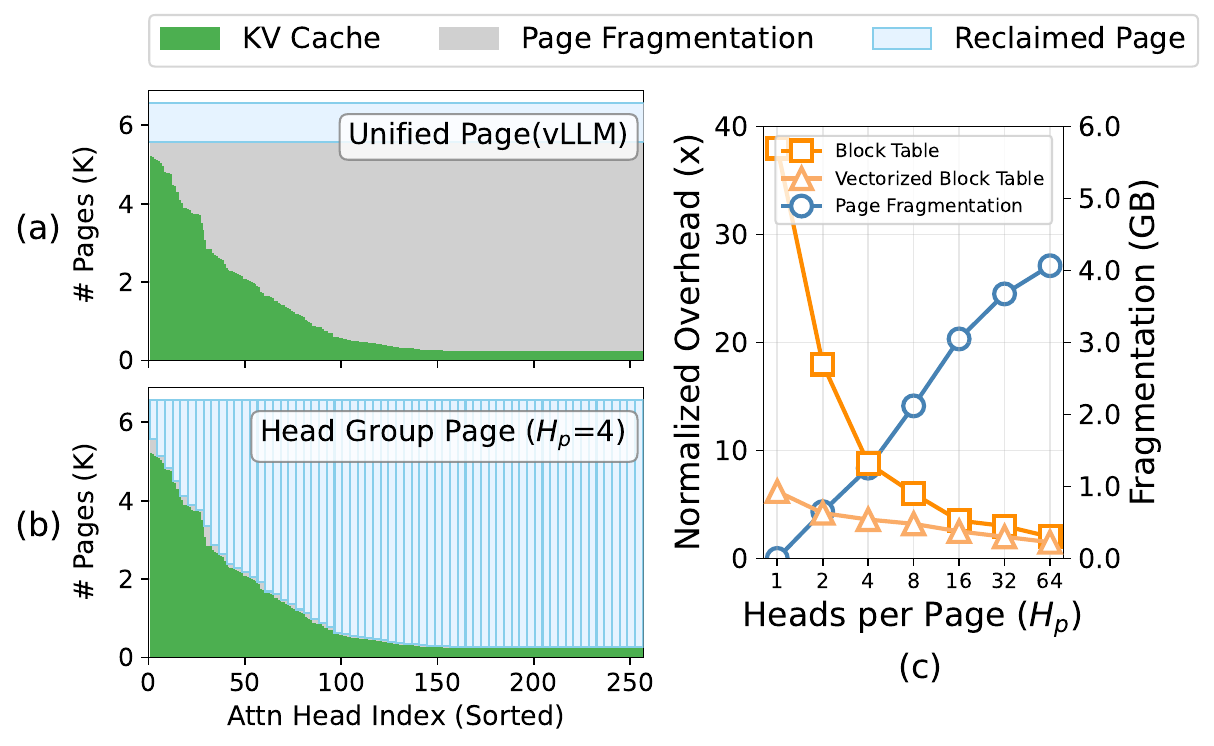}
        \caption{
        Comparison of Unified Page and Ragged Page on Llama-3.1-8B under 100K single-request non-uniform compression input at a 30\% target global retention ratio:
        (a) Unified Page (vLLM), where all heads share a single page;
        (b) Ragged Page ($H_p=4$), where each head group maintains its own independent page; and
        (c) page fragmentation versus management overhead as a function of heads per page $H_p$.
        }
    \label{fig:group_page_attn}
\end{figure}

The physical page shrinks accordingly: instead of spanning all layers and heads, a page holds $H_p \times 2 \times P \times d$ elements for the $H_p$ heads of one group in one layer, and each group owns an independent page table sized by its local maximum budget $\max_{h\in\mathcal{G}_{\ell,i}} B_{\ell,h}$ rather than the global maximum (Figure~\ref{fig:paging}(b)). The KV cache thereby becomes ragged: every head group is kept at its own \emph{effective} length---the entries it actually retains---so both a request's total footprint and its attention computation scale with the sum of per-group effective lengths (Figure~\ref{fig:system_overview}(d)). Figure~\ref{fig:group_page_attn} confirms this with a real 100K-token request: ragged paging (b) reclaims the capacity evicted from short-retention groups, which a single global group (a) keeps trapped.

\begin{algorithm}[t]
\caption{AOT Workload Partitioning with Head Group}
\label{alg:proportional_split}
\begin{algorithmic}[1]
\Require Calibrated budget-ratio tensor $\mathbf{B} \in (0,1]^{L \times H}$
         where $L$ is the number of layers and $H$ is the number of KV heads,
         available Cooperative Thread Arrays (CTAs) $N_{\mathrm{CTA}}$, heads per page $H_p$
\Ensure Static Workload Split Map $\mathbf{S} \in \mathbb{N}^{L \times (H/H_p)}$
\State $\mathbf{S} \gets \mathbf{1}_{L \times (H/H_p)}$ \Comment{initialize split factors per head group}
\For{$\ell \gets 1$ \textbf{to} $L$}
    \State $\Omega_{\ell} \gets \sum_{h=1}^{H} B_{\ell,h}$
           \Comment{total budget of layer $\ell$}
    \If{$\Omega_{\ell} = 0$} \textbf{continue} \EndIf
    \State $\tau_{\ell} \gets \Omega_{\ell} / N_{\mathrm{CTA}}$
           \Comment{target budget per split}

    \For{$i \gets 0$ \textbf{to} $H/H_p - 1$} \Comment{iterate over head groups}
        \State $\mathcal{G}_{\ell,i} \gets$ the $i$-th head group of layer $\ell$
               \Comment{grouping from \S\ref{sec:grouped_paging} (clustered or adjacent)}
        \State $\Phi_{\ell,i} \gets \sum_{h \in \mathcal{G}_{\ell,i}} B_{\ell,h}$
               \Comment{aggregated group budget}
        \State $S_{\ell,i} \gets \max\!\bigl(1,\ \mathrm{round}(\Phi_{\ell,i} / \tau_{\ell})\bigr)$
               \Comment{CTAs $\propto$ group budget share}
    \EndFor
\EndFor
\State \Return $\mathbf{S}$
\end{algorithmic}
\end{algorithm}

\subsubsection{Budget-Aware Clustering}
\label{sec:method:clustering}

The remaining question is which heads to manage together. All heads in a head-group share one page table sized to the group's largest budget, so naively grouping adjacent heads mixes dissimilar budgets and grants low-budget heads far more pages than they need. Co-locating heads of similar budgets in the same group is therefore more memory-efficient: each head-group's footprint tightly tracks what its members actually retain. Because the ranking of head budgets is model-intrinsic and input-invariant (\S\ref{key_observation}, Figure~\ref{fig:kvsize}), \textsc{Tangram} performs this \emph{Budget-Aware Clustering} once, offline.

For each layer $\ell$, we sort all $H$ heads by their calibrated budget (\S\ref{sec:fixed_budget})---that is, let $\pi_{\ell}$ be a permutation of the head indices such that the budget is monotonically increasing:
$$B_{\ell,\pi_{\ell}(0)} \leq B_{\ell,\pi_{\ell}(1)} \leq \dots \leq B_{\ell,\pi_{\ell}(H-1)}$$
The $i$-th head group is then constructed by taking $H_p$ consecutive heads from this sorted order:
$$\mathcal{G}_{\ell,i} = \{ \pi_{\ell}(j) \mid j \in [i \cdot H_p, (i+1) \cdot H_p - 1] \}$$
Figure~\ref{fig:paging}(c) shows the effect: packing short-retention heads together frees the most pages. The slack within a group is not wasted: since the shared page table holds KV up to the group's budget, every head retains at least its own budget, and smaller-budget heads use the leftover room to keep extra KV---free accuracy headroom at zero reclamation cost.

\paragraph{Balancing Memory Gain and Management Overhead.}
The number of heads per page $H_p$ governs a fundamental trade-off between memory efficiency and management overhead:
\begin{itemize}
    \item \textbf{Smaller $H_p$} tightens budget alignment within each group, enabling fine-grained memory management that maximizes reclamation---but it multiplies the number of page tables (up to $H$ when $H_p=1$), so allocation, compression, and block-table updates grow by $H/H_p$ per request and can bottleneck the host CPU.
    \item \textbf{Larger $H_p$} amortizes this management overhead across fewer page tables, but can only group heads coarsely, mixing dissimilar budgets and leaving residual fragmentation.
\end{itemize}
To balance the two, we select an appropriate $H_p$ as the operating point and introduce a \textit{Vectorized Block Table} that cuts the management overhead of maintaining multiple page tables.

\subsubsection{Vectorized Block Table Management}
\label{sec:method:vbt}

Naively, block-table cost scales with the number of groups ($\mathcal{O}(N_{req}\times H/H_p)$), making the CPU-side scheduler a bottleneck at fine-grained grouping (small $H_p$). We instead aggregate per-group block mappings into a \textit{Vectorized Block Table}, parallelizing across head groups with OpenMP and processing each group with SIMD intrinsics (e.g., AVX-512). As shown in Figure~\ref{fig:group_page_attn}(c), this shifts the fragmentation--overhead trade-off curve downward, enabling small $H_p$ to maximize memory savings without degrading end-to-end serving throughput.

\subsection{Ahead-of-Time (AOT) Load Balancing: Mitigating Workload Imbalance}
\label{sec:load_balancing}
Finally, we address the workload skew across GPU SMs in decode attention (\S~\ref{sec:workload_imbalance}). Because the per-head footprint is static after calibration, the computational load is fully predictable, allowing the entire load-balancing burden to move off the critical path.

\begin{figure*}[t]
    \centering
    \includegraphics[width=\linewidth]{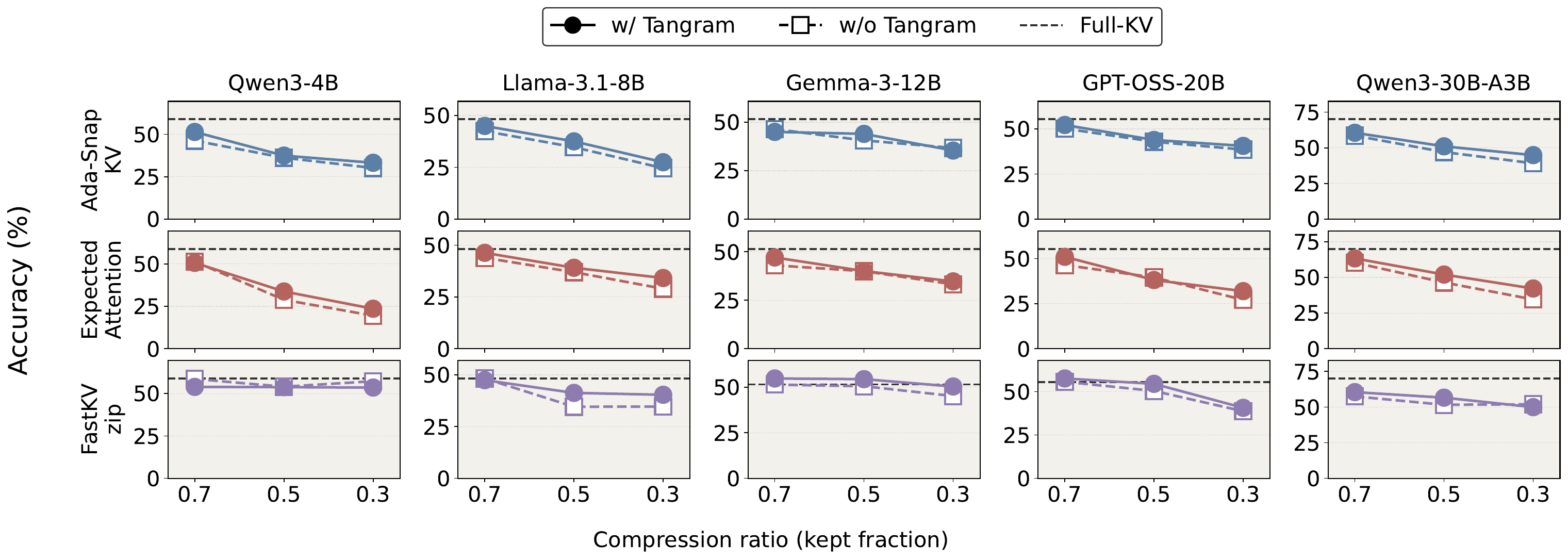}
    \caption{Multi-turn accuracy of three non-uniform KV compression methods (rows) across five models (columns), at target global retention ratios $\rho$ of 0.7/0.5/0.3 and averaged over all SCBench tasks. For each method, \emph{w/o Tangram} (dashed, open markers) is its original implementation, while \emph{w/ Tangram} (solid, filled markers) is the same method run under \textsc{Tangram}'s Ragged Paging and budget reservation. The horizontal dashed line denotes the Full-KV reference.}
    \label{fig:acc}
\end{figure*}

\subsubsection{Ahead-of-Time (AOT) Workload Split Map}
\label{sec:aot_split}
As shown in Figure~\ref{fig:system_overview}(e), since the per-head budget $B_{\ell,h}$ is determined offline and remains constant across requests, the ``shape'' of the computation is known before inference. We pre-calculate a static \textit{Workload Split Map} $\mathbf{S} \in \mathbb{N}^{L \times (H/H_p)}$ to enforce perfectly balanced parallelism.

\paragraph{Static Partitioning Strategy.}

To maximize hardware utilization, we first leverage the \texttt{CUDAMaxOccupancy} API to determine the total number of these CTAs, denoted as $N_{\mathrm{CTA}}$, that the target GPU can execute concurrently for the attention kernel. This value represents the device's aggregate parallelism capacity. We then employ an \textit{Ahead-of-Time (AOT) Workload Partitioning} algorithm (Algorithm~\ref{alg:proportional_split}) to distribute these CTAs across head groups proportional to their aggregated computational weight. Specifically, each head group's budget $\Phi_{\ell,i}$ is computed by summing the calibrated budget ratios of all heads within the group. Head groups with large aggregated budgets are assigned higher partition factors, while groups with small aggregated budgets are assigned fewer partitions. The output is stored in the static Workload Split Map $\mathbf{S}$, where each entry $S_{\ell,i}$ dictates exactly how many thread blocks should be allocated for head group $i$ in layer $\ell$. This ensures that the total work assigned to each CTA is approximately equal, thereby eliminating tail latency in which the entire system stalls while waiting for a single overloaded CTA to complete.

\paragraph{Runtime Execution} During decoding, \textsc{Tangram} simply retrieves the precomputed table $\mathbf{S}$ to configure the attention kernel, incurring zero runtime planning cost. The static plan remains valid because $\mathbf{S}$ is computed from budget
\emph{ratios}, not absolute lengths: every head group's retained length scales linearly with the same context length $N$ (i.e., $\lceil B_{\ell,h}N\rceil$), so the relative workload across
groups---the only quantity that determines balance---is invariant to $N$, and batching preserves it since each request contributes work in the same fixed proportions. The plan thus stays near-optimal across context lengths and batch sizes---precisely what per-layer dynamic
planning~\cite{ye2025flashinfer} spends 15--20\% of every decode step rediscovering (Table~\ref{tab:load_balancing_overhead}).

\section{Evaluation} 

\subsection{Evaluation Setup}
\label{sec:eval-setup}

\paragraph{Models and Workloads.} We evaluate \textsc{Tangram} on five models spanning dense and Mixture-of-Experts architectures — Qwen3-4B, Llama-3.1-8B, Gemma-3-12B, GPT-OSS-20B, and Qwen3-30B-A3B—each supporting context windows exceeding 100K tokens, which is necessary for capturing the massive context accumulation that arises in multi-turn LLM serving. 
We adopt SCBench~\cite{li2024scbench}, which evaluates long-context capability through shared-context, multi-turn interactions spanning retrieval, reasoning, summarization, and code understanding---well-suited for stress-testing both the accuracy and efficiency of KV cache management under realistic serving conditions.
For \textsc{Tangram}'s budget reservation, we utilize pre-determined budgets derived offline from 50 pilot samples, setting the safety coefficient $\alpha$ to 2 across all evaluations. Throughout, we report the target global retention ratio $\rho$ as a percentage, where $\rho{=}100\%$ denotes the
uncompressed Full-KV cache.

To systematically quantify the performance gains of our framework across varying context scales, we partition the evaluation tasks into three categories: (1)~\textbf{Short} (< 20K tokens), represented by the Many-Shot task; (2)~\textbf{Mid} (20K--100K tokens), covering RepoQA, Multi-Choice QA, and MathFind tasks that exercise moderate-to-heavy context accumulation and require selective retrieval over substantial histories; and (3)~\textbf{Long} (> 100K tokens), encompassing Retrieve Prefix-Suffix, KV, and Summary, where the KV cache footprint of even a few requests dominates GPU memory. Together, these workloads comprehensively evaluate the scalability and efficiency of \textsc{Tangram} across the full spectrum of context lengths encountered in multi-turn LLM serving.

\begin{figure}[t]
    \centering
    \includegraphics[width=\linewidth]{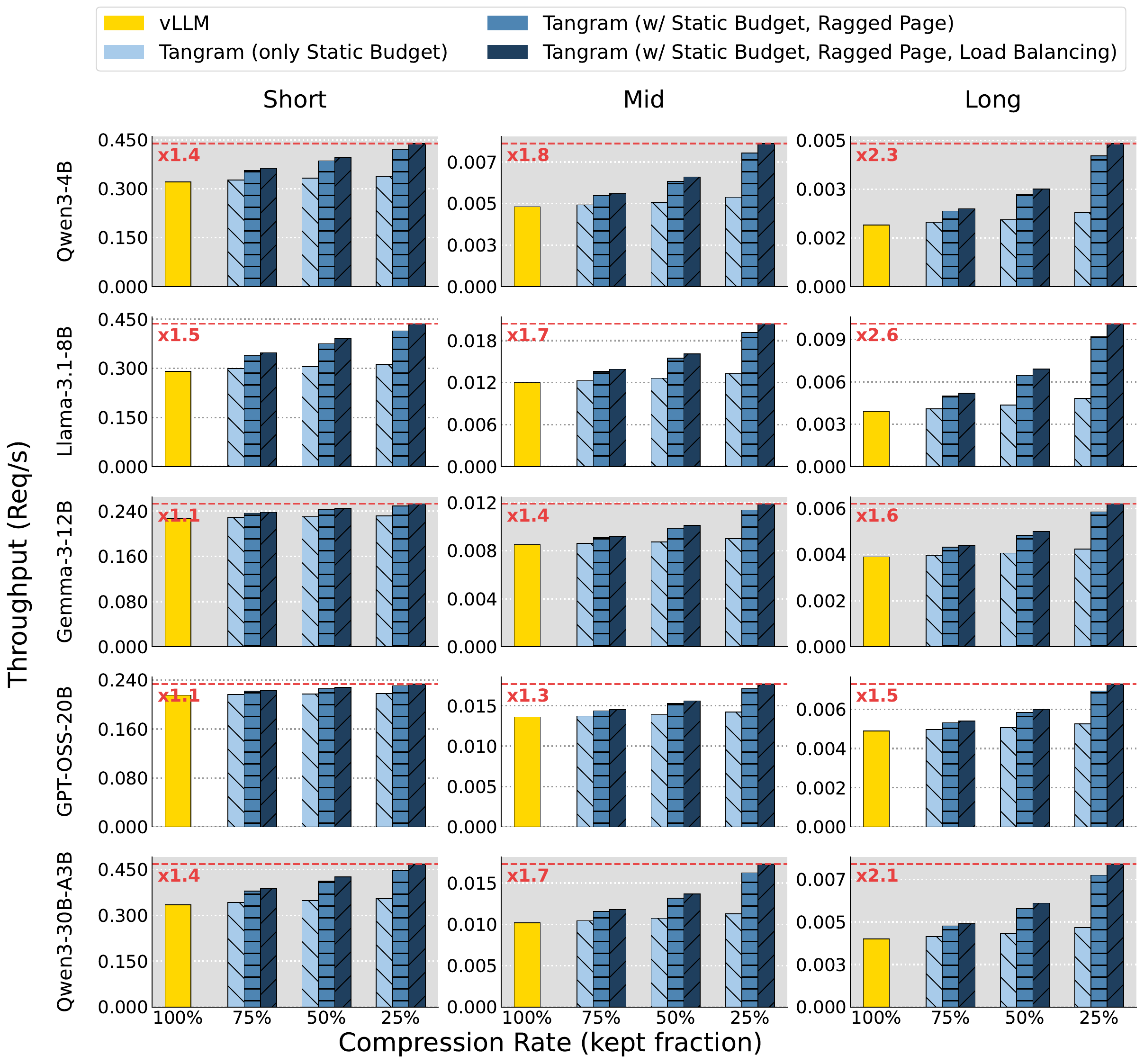}
    \caption{Throughput breakdown on SCBench~\cite{li2024scbench} across target global retention ratios ($H_p=4$). Against the vLLM baseline ($\rho=100\%$), \textsc{Tangram}'s techniques are applied cumulatively: Budget Reservation, then Ragged Paging, then AOT Load Balancing. The red ×N marks \textsc{Tangram}'s peak throughput ($\rho=25\%$) over vLLM.}
    \label{fig:Throughput_breakdown}
\end{figure}

\paragraph{System Setup.} We implement \textsc{Tangram} on top of vLLM~\cite{kwon2023efficient}, a state-of-the-art high-throughput serving framework. To support our proposed non-uniform KV cache compression, we integrate specialized CUDA kernels developed based on FlashAttention~\cite{dao2023flashattention}, ensuring our custom operators remain fully compatible with standard attention interfaces~\footnote{Our customized vLLM implementation remains fully compatible with the open-source frameworks and will be released to the community to accelerate innovation.}. All end-to-end experiments are conducted on a dedicated server node equipped with an Intel(R) Xeon(R) Gold 6326 CPU @ 2.90 GHz (16 physical cores) and four NVIDIA A100 GPUs (80GB of memory each). We emphasize that all reported results—including throughput, latency, and fragmentation rates—are empirical measurements obtained from actual runtime execution on this hardware, rather than analytical estimates or simulations.

\paragraph{Baselines.} For the non-uniform compression methods, we use their original PyTorch implementations, which dynamically allocate per-head budgets at runtime to meet a target global retention ratio; these provide the \emph{w/o Tangram} references in \S\ref{sec:ltc_acc_eval}. For load balancing, we compare against two state-of-the-art attention kernels. FlashDecoding~\cite{dao_flashdecoding} parallelizes attention with a \emph{Split-KV} strategy; effective for uniform caches, its static, heuristic partitioning suffers severe stragglers under the workload skew of non-uniform caches. FlashInfer~\cite{ye2025flashinfer} instead plans partitions at run time, but recomputing a unique plan for every heterogeneous layer incurs significant CPU latency (15--20\% of decoding time), negating its GPU parallelism.

\begin{figure}[t]
    \centering
    \includegraphics[width=\linewidth]{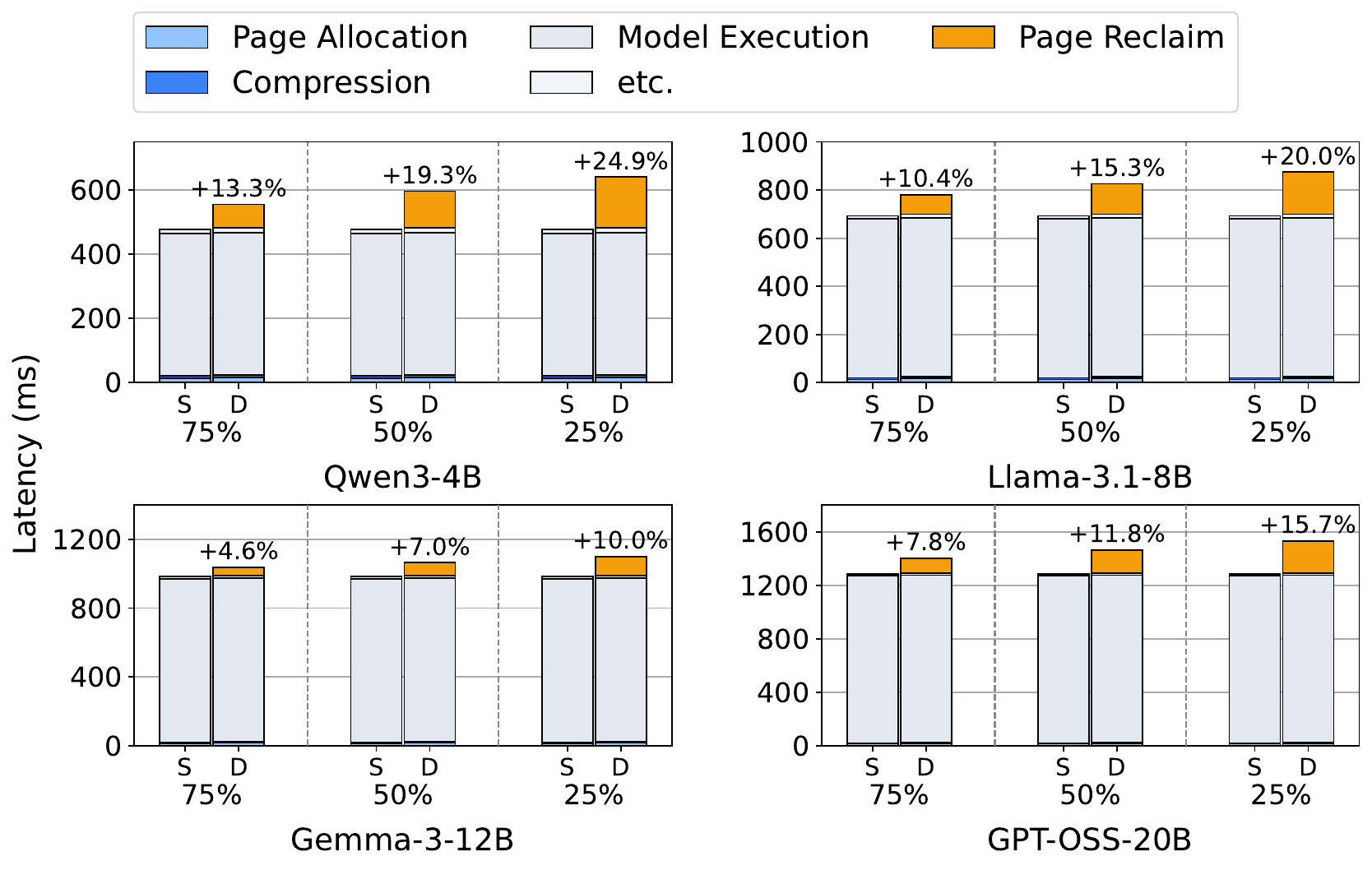}
    \caption{Latency breakdown across various target global retention ratios, comparing static budget allocation (\emph{S}) against dynamic allocation (\emph{D}). While dynamic allocation incurs significant page reclamation overhead (orange) that scales with the eviction rate, \textsc{Tangram}'s budget reservation completely eliminates this extra cost, operating without page reclaim overhead.}
    \label{fig:latency_breakdown}
\end{figure}

\begin{figure}[t]
    \centering
    \setlength{\abovecaptionskip}{2pt}  %
    \includegraphics[width=\linewidth]{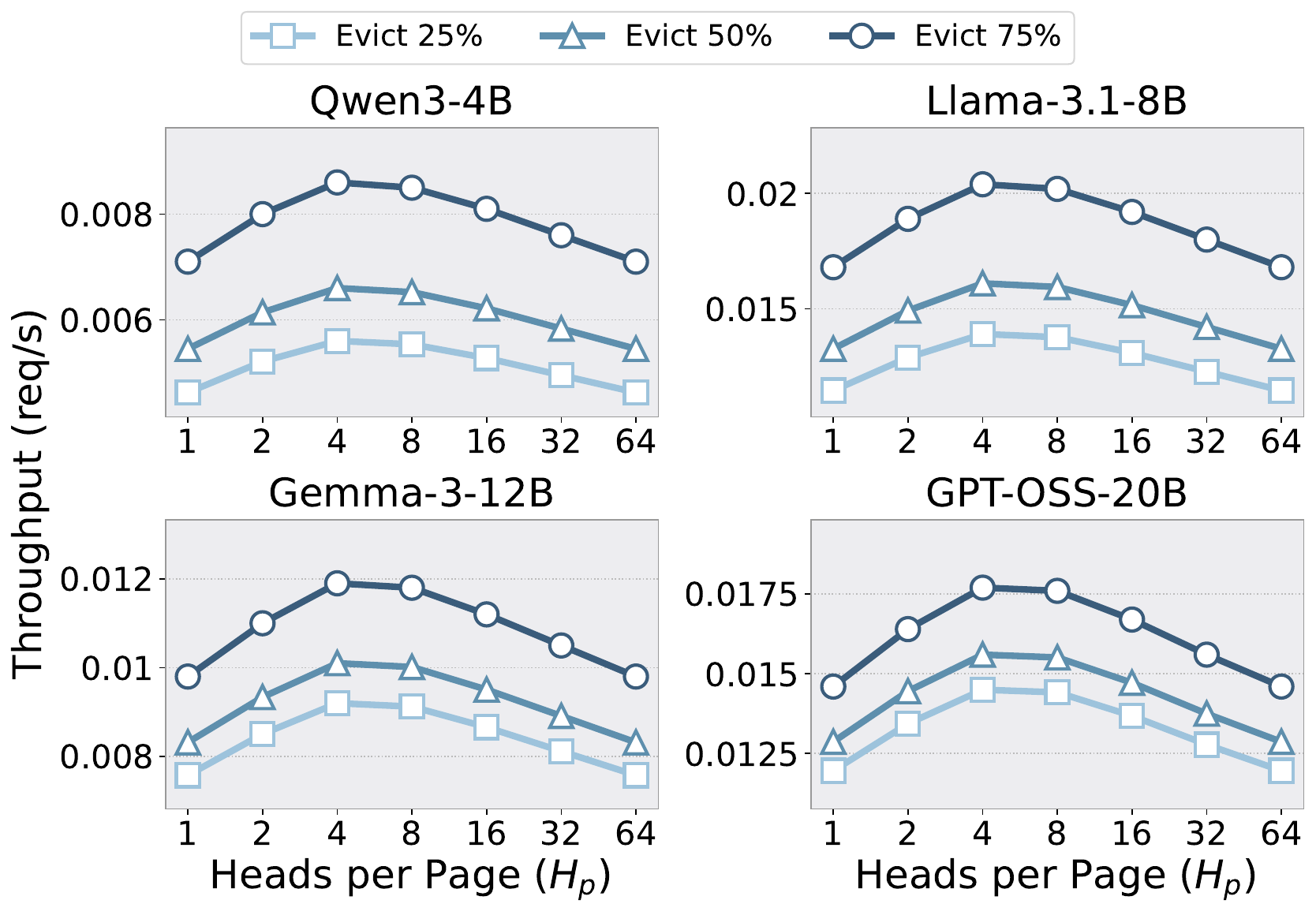}
    \caption{Throughput versus heads per page ($H_p$); the three curves correspond to eviction rates of 25/50/75\%. Small $H_p$ minimizes fragmentation but incurs high management overhead, while large $H_p$ fails to effectively reclaim memory.}
    \label{fig:head_group_wise_throughput}
\end{figure}

\paragraph{Performance Metrics.} Our evaluation relies on a comprehensive set of metrics covering both model quality and system efficiency. For Multi-turn LLM Serving capability, we report the average score across Short, Mid, and Long categories for each model, based on the accuracy metrics provided by the benchmark. For system performance, we focus on four key indicators: (1) \textit{Throughput} (requests per second), which demonstrates overall system capacity under varying load conditions and isolates the contribution of each proposed technique to the end-to-end throughput improvement; (2) \textit{Prefill Latency Breakdown} to analyze the efficiency gains from Budget Reservation; (3) \textit{Decode Attention Latency} to validate the effectiveness of AOT Load Balancing; and (4) \textit{TTFT} to verify serving performance under realistic multi-turn scenarios.

\subsection{Multi-turn Accuracy Evaluation}
\label{sec:ltc_acc_eval}
We first evaluate how \textsc{Tangram}'s memory management affects accuracy. For each non-uniform compression method, we compare its original implementation (\emph{w/o Tangram}) against the same method run under \textsc{Tangram} (\emph{w/ Tangram}), isolating the accuracy impact of Ragged Paging and budget reservation while holding the compression method fixed. As summarized in Figure~\ref{fig:acc}, across various non-uniform compression methods---Ada-SnapKV (Ada-KV's~\cite{feng2024ada} non-uniform budget allocation over SnapKV's~\cite{li2024snapkv} importance scores), Expected Attention~\cite{devoto2025expected}, and FastKVzip~\cite{kim2026fast}---and five models, \emph{w/ Tangram} closely tracks each method's original \emph{w/o Tangram} accuracy across target global retention ratios. Notably, although \textsc{Tangram} pins each head group to a static, offline-calibrated budget, it matches, and in some cases exceeds, their original accuracy, all while running free of the memory inefficiencies that previously made these methods impractical to serve.
This fidelity is by construction: each head's calibrated budget
tracks the retention that the underlying method would assign at runtime,
with the safety margin $\alpha$ absorbing per-input deviation
(\S\ref{sec:calibration}). \textsc{Tangram} thus acts as a faithful systems substrate
for non-uniform compression rather than altering the compression
itself.

\begin{figure}[t]
    \centering
    \setlength{\abovecaptionskip}{2pt}  %
    \includegraphics[width=\linewidth]{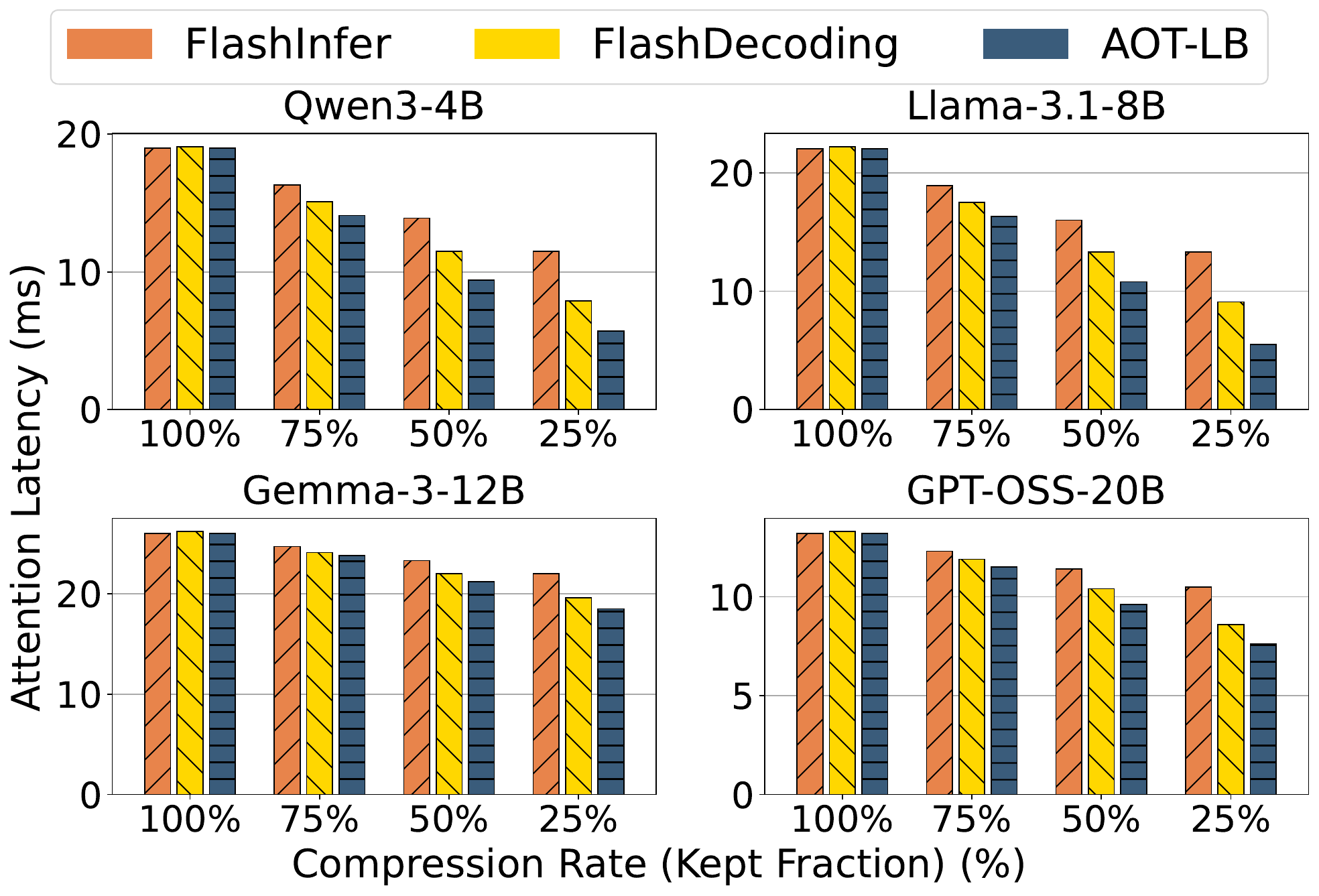}
    \caption{Attention latency evaluated under the impact of AOT (Ahead-Of-Time) load balancing (fixed batch size of 4).}
    \label{fig:attn_latency}
\end{figure}

\subsection{End-to-end Performance}
\label{sec:eval-perf}
\begin{figure}[t]
    \centering
    \setlength{\abovecaptionskip}{2pt}  %
    \includegraphics[width=\linewidth]{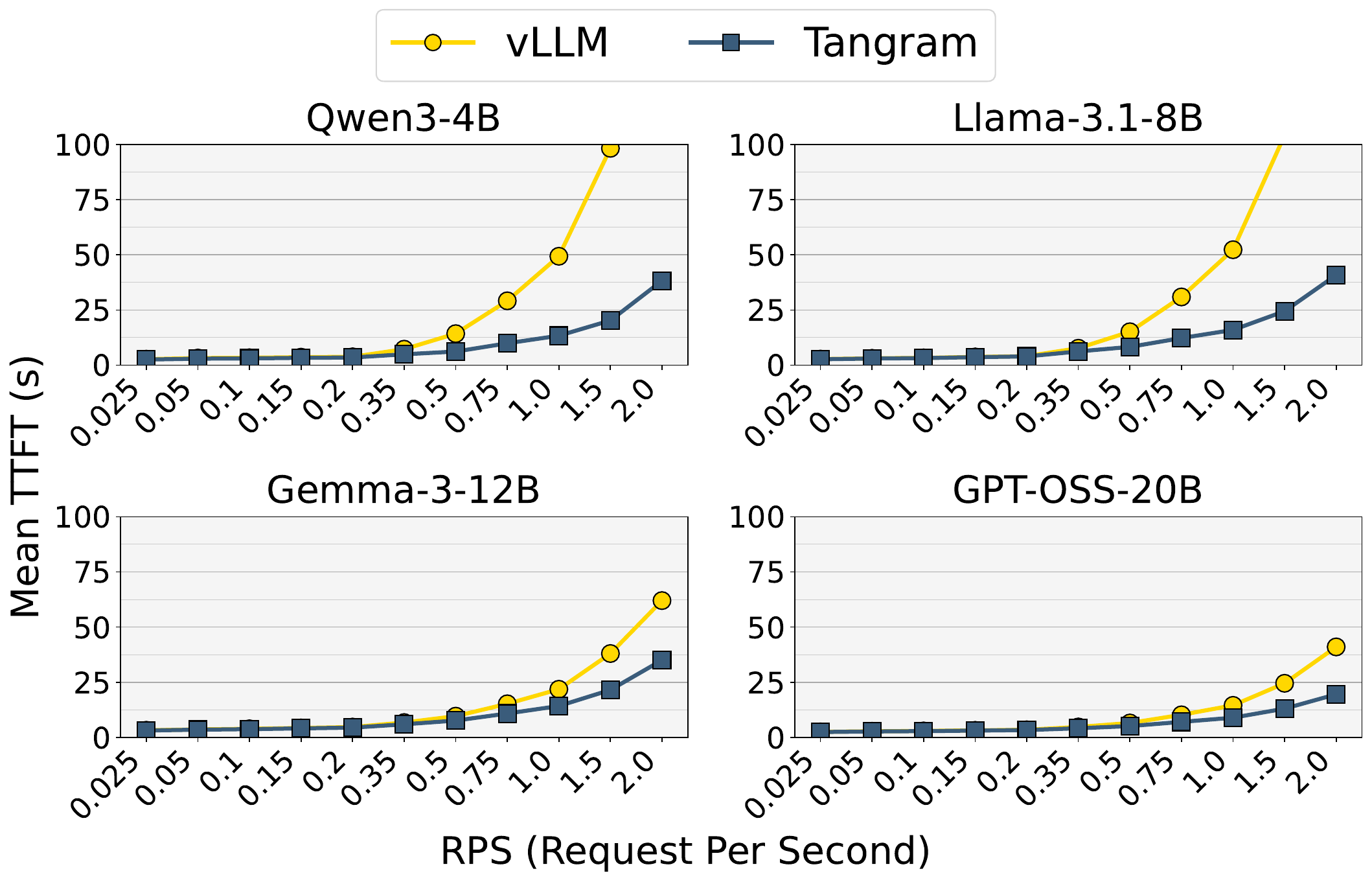}
    \caption{TTFT (Time-To-First-Token) under increasing throughput pressure with 30K average request lengths is maintained through budget reservation and Ragged Paging with 75\% of the KV cache evicted.}
    \label{fig:ttft}
\end{figure}

We evaluate the serving throughput of \textsc{Tangram} against the vLLM baseline, using FastKVzip~\cite{kim2026fast} as the state-of-the-art non-uniform compression method. As shown in Figure~\ref{fig:Throughput_breakdown}, \textsc{Tangram} successfully translates non-uniform KV compression into practical system-level gains, achieving up to a $2.6\times$ throughput improvement, with the gain growing as context length increases from Short to Long. This capacity gain also translates into better latency under heavy load: under 75\% eviction, \textsc{Tangram} sustains low TTFT as the request rate grows, whereas vLLM's TTFT rises sharply (Figure~\ref{fig:ttft}). To isolate the contribution of each proposed technique, we incrementally apply Budget Reservation(\S~\ref{sec:fixed_budget}), Ragged Paging(\S~\ref{sec:grouped_paging}), and AOT Load Balancing(\S~\ref{sec:load_balancing}). The results confirm that each component provides additive throughput gains, collectively bridging the gap between theoretical KV cache reduction and realized system performance.

We note that the $2.6\times$ gain over the vLLM Full-KV baseline combines two effects: the capacity freed by compression itself and the system efficiency contributed by \textsc{Tangram}. The cumulative ablation in Figure~\ref{fig:Throughput_breakdown} isolates the latter---moving from Budget Reservation alone to the full system---while Figure~\ref{fig:latency_breakdown} quantifies the cost a dynamic implementation would pay instead. \textsc{Tangram}'s contribution is thus not the compression ratio, which is inherited from the underlying method, but the conversion of that ratio into realized throughput.

\paragraph{Eliminating Page Reclamation Overhead.}
As shown in Figure~\ref{fig:latency_breakdown}, dynamic compression imposes severe overhead, with page reclamation consuming up to 25\% of prefill execution time to track and reclaim scattered pages. In contrast, \textsc{Tangram} incurs little extra cost. Because Budget Reservation defines the exact memory footprint before execution, our system allocates only the required pages from the outset, completely eliminating the need to perform any page reclamation.

\paragraph{Impact of Ragged Paging.} Ragged Paging is the key mechanism that translates non-uniform compression into actual memory savings by enabling independent page reclamation at the head group level. However, as discussed in \S\ref{sec:grouped_paging}, the choice of heads per page $H_p$ introduces a fundamental trade-off: excessively small $H_p$ proliferates the number of page tables, increasing management overhead that degrades system performance, while excessively large $H_p$ prevents the system from reclaiming evicted pages, as the group's allocation remains dictated by its longest-retaining head. As shown in Figure~\ref{fig:head_group_wise_throughput}, we observe that $H_p=4$--$8$ strikes the optimal balance, yielding the highest end-to-end throughput across all configurations.

\begin{figure}[t]
    \centering
    \setlength{\abovecaptionskip}{2pt}  %
    \includegraphics[width=\linewidth]{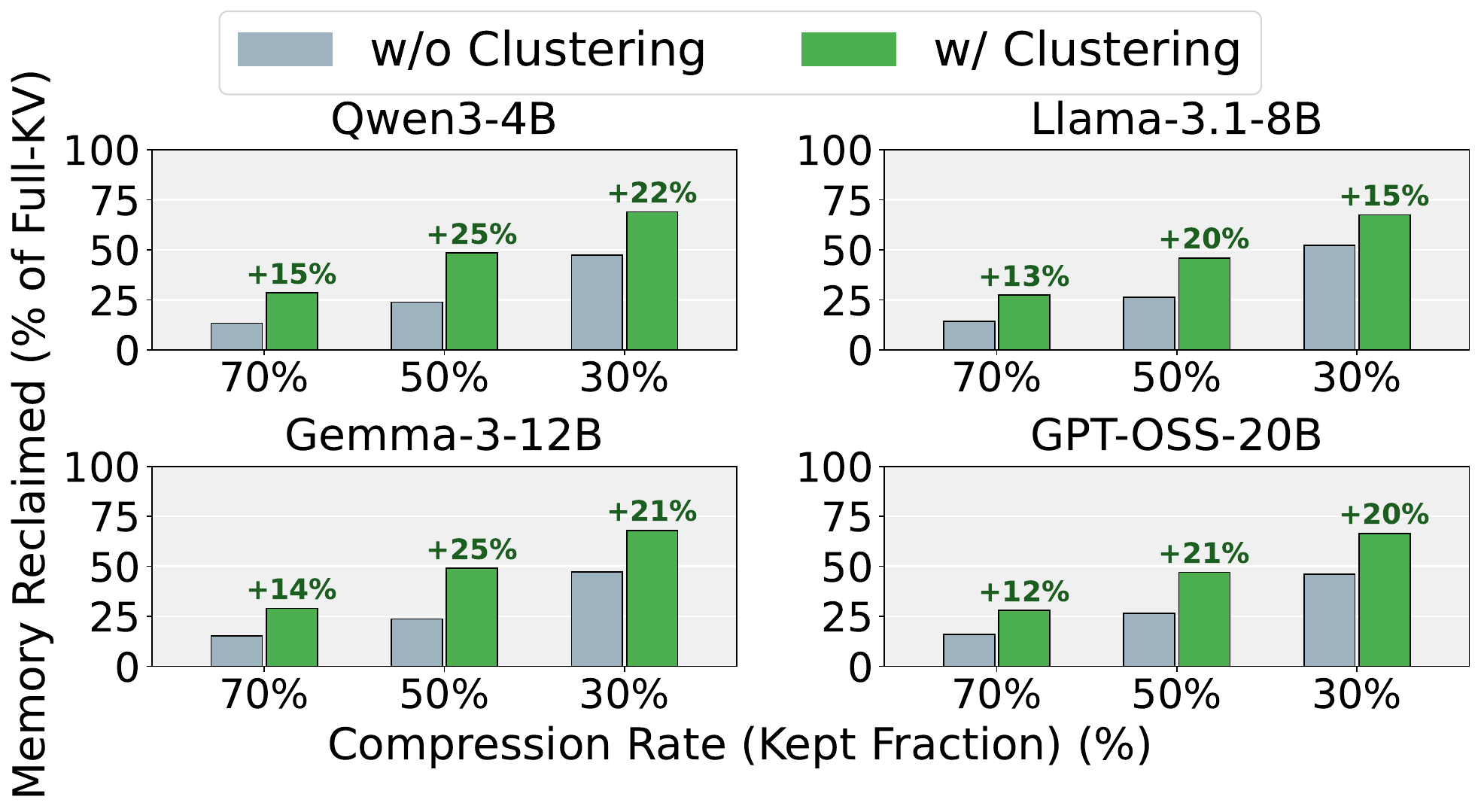}
    \caption{Memory reclaimed (fraction of the full KV cache) with and without Budget-Aware Clustering ($H_p=4$).}
    \label{fig:clustering_reclaim}
\end{figure}

\paragraph{Effectiveness of Budget-Aware Clustering.} Beyond the group size, \emph{which} heads share a page table is equally decisive: grouping dissimilar heads inflates each group's allocation to its largest member and leaves the surplus unreclaimable. Budget-Aware Clustering removes this slack by co-locating heads of similar budget, so each group's footprint tightly tracks the retention its members actually need. As shown in Figure~\ref{fig:clustering_reclaim}, clustering reclaims an additional 12--25\% of the full KV cache over grouping adjacent heads at the same $H_p$, consistently across diverse models.

\paragraph{Efficient Load Balancing.} As shown in Figure~\ref{fig:attn_latency}, our AOT Load Balancing consistently achieves the lowest decode attention latency. FlashDecoding suffers from straggler effects due to its heuristic static partitioning, while FlashInfer incurs significant overhead from recomputing per-layer partitions at every decoding step. \textsc{Tangram} avoids both issues by pre-calculating optimal workload partitions offline, achieving balanced SM utilization without runtime cost.

\section{Related Works}\label{sec:related_works}

\paragraph{Multi-turn LLM Serving.} As LLMs evolve into persistent assistants, maintaining user-specific context across sessions has become critical~\cite{achiam2023gpt,reid2024gemini,openai2024memory,anthropic2024claudememory}. Recent benchmarks like SCBench~\cite{li2024scbench}, RealTalk~\cite{lee2025realtalk} and LoCoMo~\cite{maharana-etal-2024-evaluating} highlight the difficulty of recalling long-horizon details, motivating algorithmic solutions such as retrieval augmentation~\cite{zhong2024memorybank,chhikara2025mem0,kim2025epicacheepisodickvcache}. However, prior work largely overlooks the serving efficiency of these memory-intensive workloads. Our work bridges this gap by addressing the system bottlenecks of managing the rapidly scaling KV cache required for robust long-term memory.

\paragraph{KV Cache compression.} Compression techniques are essential for reducing memory pressure. \emph{Uniform} compression enforces uniform retention across all attention heads~\cite{jiang2023mistral7b,xiao2024efficient,zhang2023ho,oren-etal-2024-transformers,li2024snapkv,kim-etal-2024-infinipot}, simplifying management but often discarding context essential for specific heads. In contrast, \emph{non-uniform} compression improves accuracy by allowing heterogeneous retention budgets~\cite{feng2024ada,kim2025kvzip,xiao2025duoattention,fu2025not,tang2025razorattention}. While algorithmically superior, non-uniform methods have been impractical for deployment due to system-level incompatibilities. We identify and resolve the core barriers—fragmentation, scheduling uncertainty, and workload imbalance—to make non-uniform compression viable in production.

\paragraph{Heterogeneous Memory Management.} Recent systems have begun
to accommodate KV heterogeneity: Jenga~\cite{zhang2025jenga} manages it across layer
types in hybrid models~\cite{dong2024hymba,lieber2024jamba,team2024gemma} but does not target the explosive
multi-turn KV growth, while DiffKV~\cite{zhang2025diffkv} compresses caches with
sparsity and quantization but manages each head independently.
Neither exploits the cross-head structural regularity that
\textsc{Tangram} identifies; by clustering heads of similar,
model-intrinsic retention into ragged pages, \textsc{Tangram} aligns
page boundaries with the actual retention distribution, making
head-level heterogeneous memory management practical for
high-throughput serving.

\section{Conclusion}
We present \textsc{Tangram}, a serving system that brings non-uniform
KV cache compression to real LLM serving stacks. Exploiting the
model-intrinsic regularity of head-wise retention, \textsc{Tangram}
statically resolves what prior systems handle at runtime---reserving
exact post-compression budgets at scheduling time, reclaiming
fragmented memory through budget-clustered ragged pages, and
balancing GPU workloads with a precomputed Workload Split Map. Together,
these make non-uniform compression practical, delivering up to
2.6$\times$ higher throughput while matching the accuracy of the
underlying non-uniform compression methods.

\bibliographystyle{ACM-Reference-Format}
\bibliography{refs}

@inproceedings{kwon2023efficient,
  title={Efficient memory management for large language model serving with pagedattention},
  author={Kwon, Woosuk and Li, Zhuohan and Zhuang, Siyuan and Sheng, Ying and Zheng, Lianmin and Yu, Cody Hao and Gonzalez, Joseph and Zhang, Hao and Stoica, Ion},
  booktitle={Proceedings of the 29th symposium on operating systems principles},
  pages={611--626},
  year={2023}
}

@article{kim2026fast,
  title={Fast KVzip: Efficient and Accurate LLM Inference with Gated KV Eviction},
  author={Kim, Jang-Hyun and Han, Dongyoon and Yun, Sangdoo},
  journal={arXiv preprint arXiv:2601.17668},
  year={2026}
}

@inproceedings{zhang2025diffkv,
  title={DiffKV: Differentiated Memory Management for Large Language Models with Parallel KV Compaction},
  author={Zhang, Yanqi and Hu, Yuwei and Zhao, Runyuan and Lui, John CS and Chen, Haibo},
  booktitle={Proceedings of the ACM SIGOPS 31st Symposium on Operating Systems Principles},
  pages={431--445},
  year={2025}
}

@article{zheng2023efficiently,
  title={Efficiently Programming Large Language Models using SGLang.},
  author={Zheng, Lianmin and Yin, Liangsheng and Xie, Zhiqiang and Huang, Jeff and Sun, Chuyue and Yu, Cody\_Hao and Cao, Shiyi and Kozyrakis, Christos and Stoica, Ion and Gonzalez, Joseph E and others},
  year={2023},
  journal={arXiv preprint arXiv:2312.07104},
  publisher={arXiv}
}

@article{dao2023flashattention,
  title={Flashattention-2: Faster attention with better parallelism and work partitioning},
  author={Dao, Tri},
  journal={arXiv preprint arXiv:2307.08691},
  year={2023}
}

@inproceedings{maharana-etal-2024-evaluating,
    title = "Evaluating Very Long-Term Conversational Memory of {LLM} Agents",
    author = "Maharana, Adyasha  and
      Lee, Dong-Ho  and
      Tulyakov, Sergey  and
      Bansal, Mohit  and
      Barbieri, Francesco  and
      Fang, Yuwei",
    editor = "Ku, Lun-Wei  and
      Martins, Andre  and
      Srikumar, Vivek",
    booktitle = "Proceedings of the 62nd Annual Meeting of the Association for Computational Linguistics (Volume 1: Long Papers)",
    month = aug,
    year = "2024",
    address = "Bangkok, Thailand",
    publisher = "Association for Computational Linguistics",
    url = "https://aclanthology.org/2024.acl-long.747/",
    doi = "10.18653/v1/2024.acl-long.747",
    pages = "13851--13870",
}

@article{lee2025realtalk,
  title={Realtalk: A 21-day real-world dataset for long-term conversation},
  author={Lee, Dong-Ho and Maharana, Adyasha and Pujara, Jay and Ren, Xiang and Barbieri, Francesco},
  journal={arXiv preprint arXiv:2502.13270},
  year={2025}
}

@inproceedings{
wu2025longmemeval,
title={LongMemEval: Benchmarking Chat Assistants on Long-Term Interactive Memory},
author={Di Wu and Hongwei Wang and Wenhao Yu and Yuwei Zhang and Kai-Wei Chang and Dong Yu},
booktitle={The Thirteenth International Conference on Learning Representations},
year={2025},
url={https://openreview.net/forum?id=pZiyCaVuti}
}

@article{pope2023efficiently,
  title={Efficiently scaling transformer inference},
  author={Pope, Reiner and Douglas, Sholto and Chowdhery, Aakanksha and Devlin, Jacob and Bradbury, James and Heek, Jonathan and Xiao, Kefan and Agrawal, Shivani and Dean, Jeff},
  journal={Proceedings of machine learning and systems},
  volume={5},
  pages={606--624},
  year={2023}
}

@inproceedings{zhong2024distserve,
  title={$\{$DistServe$\}$: Disaggregating prefill and decoding for goodput-optimized large language model serving},
  author={Zhong, Yinmin and Liu, Shengyu and Chen, Junda and Hu, Jianbo and Zhu, Yibo and Liu, Xuanzhe and Jin, Xin and Zhang, Hao},
  booktitle={18th USENIX Symposium on Operating Systems Design and Implementation (OSDI 24)},
  pages={193--210},
  year={2024}
}

@article{ye2025flashinfer,
  title={Flashinfer: Efficient and customizable attention engine for llm inference serving},
  author={Ye, Zihao and Chen, Lequn and Lai, Ruihang and Lin, Wuwei and Zhang, Yineng and Wang, Stephanie and Chen, Tianqi and Kasikci, Baris and Grover, Vinod and Krishnamurthy, Arvind and others},
  journal={arXiv preprint arXiv:2501.01005},
  year={2025}
}

@article{li2024scbench,
  title={Scbench: A kv cache-centric analysis of long-context methods},
  author={Li, Yucheng and Jiang, Huiqiang and Wu, Qianhui and Luo, Xufang and Ahn, Surin and Zhang, Chengruidong and Abdi, Amir H and Li, Dongsheng and Gao, Jianfeng and Yang, Yuqing and others},
  journal={arXiv preprint arXiv:2412.10319},
  year={2024}
}

@misc{dao_flashdecoding,
  title        = {{Flash-Decoding for long-context inference}},
  author       = {Dao, Tri and Haziza, Daniel and Massa, Francisco and Sizov, Grigory},
  howpublished = {\url{https://crfm.stanford.edu/2023/10/12/flashdecoding.html}},
  year         = {2023},
  note         = {}
}

@misc{cudamaxoccupancy,
  title        = {CUDA Max Occupancy API},
  author       = {NVIDIA},
  howpublished = {https://docs.nvidia.com/cuda/cuda-runtime-api/group__CUDART__OCCUPANCY.html},
  year         = {},
  note         = {}
}

@article{achiam2023gpt,
  title={Gpt-4 technical report},
  author={Achiam, Josh and Adler, Steven and Agarwal, Sandhini and Ahmad, Lama and Akkaya, Ilge and Aleman, Florencia Leoni and Almeida, Diogo and Altenschmidt, Janko and Altman, Sam and Anadkat, Shyamal and others},
  journal={arXiv preprint arXiv:2303.08774},
  year={2023}
}

@inproceedings{
li2024snapkv,
title={Snap{KV}: {LLM} Knows What You are Looking for Before Generation},
author={Yuhong Li and Yingbing Huang and Bowen Yang and Bharat Venkitesh and Acyr Locatelli and Hanchen Ye and Tianle Cai and Patrick Lewis and Deming Chen},
booktitle={The Thirty-eighth Annual Conference on Neural Information Processing Systems},
year={2024},
url={https://openreview.net/forum?id=poE54GOq2l}
}

@inproceedings{
zhang2023ho,
title={H2O: Heavy-Hitter Oracle for Efficient Generative Inference of Large Language Models},
author={Zhenyu Zhang and Ying Sheng and Tianyi Zhou and Tianlong Chen and Lianmin Zheng and Ruisi Cai and Zhao Song and Yuandong Tian and Christopher Re and Clark Barrett and Zhangyang Wang and Beidi Chen},
booktitle={Thirty-seventh Conference on Neural Information Processing Systems},
year={2023},
url={https://openreview.net/forum?id=RkRrPp7GKO}
}

@article{kim2025kvzip,
        title={KVzip: Query-Agnostic KV Cache Compression with Context Reconstruction},
        author={Kim, Jang-Hyun and Kim, Jinuk and Kwon, Sangwoo and Lee, Jae W and Yun, Sangdoo and Song, Hyun Oh},
        journal={Advances in Neural Information Processing Systems},
        year={2025}
}

@article{feng2024ada,
  title={Ada-kv: Optimizing kv cache eviction by adaptive budget allocation for efficient llm inference},
  author={Feng, Yuan and Lv, Junlin and Cao, Yukun and Xie, Xike and Zhou, S Kevin},
  journal={arXiv preprint arXiv:2407.11550},
  year={2024}
}

@misc{openai2024memory,
  author       = {OpenAI},
  title        = {Memory and New Controls for ChatGPT},
  year         = {2024},
  howpublished = {\url{https://openai.com/index/memory-and-new-controls-for-chatgpt/}},
}

@misc{anthropic2024claudememory,
  author       = {Anthropic},
  title        = {Using Claude's Chat, Search, and Memory to Build on Previous Context},
  year         = {2024},
  howpublished = {\url{https://support.claude.com/en/articles/11817273}},
}

@article{reid2024gemini,
  title={Gemini 1.5: Unlocking multimodal understanding across millions of tokens of context},
  author={Reid, Machel and Savinov, Nikolay and Teplyashin, Denis and Lepikhin, Dmitry and Lillicrap, Timothy and Alayrac, Jean-baptiste and Soricut, Radu and Lazaridou, Angeliki and Firat, Orhan and Schrittwieser, Julian and others},
  journal={arXiv preprint arXiv:2403.05530},
  year={2024}
}

@inproceedings{kim-etal-2024-infinipot,
    title = "{I}nfini{P}ot: Infinite Context Processing on Memory-Constrained {LLM}s",
    author = "Kim, Minsoo  and
      Shim, Kyuhong  and
      Choi, Jungwook  and
      Chang, Simyung",
    editor = "Al-Onaizan, Yaser  and
      Bansal, Mohit  and
      Chen, Yun-Nung",
    booktitle = "Proceedings of the 2024 Conference on Empirical Methods in Natural Language Processing",
    month = nov,
    year = "2024",
    address = "Miami, Florida, USA",
    publisher = "Association for Computational Linguistics",
    url = "https://aclanthology.org/2024.emnlp-main.897/",
    doi = "10.18653/v1/2024.emnlp-main.897",
    pages = "16046--16060"
}

@misc{jiang2023mistral7b,
      title={Mistral 7B}, 
      author={Albert Q. Jiang and Alexandre Sablayrolles and Arthur Mensch and Chris Bamford and Devendra Singh Chaplot and Diego de las Casas and Florian Bressand and Gianna Lengyel and Guillaume Lample and Lucile Saulnier and Lélio Renard Lavaud and Marie-Anne Lachaux and Pierre Stock and Teven Le Scao and Thibaut Lavril and Thomas Wang and Timothée Lacroix and William El Sayed},
      year={2023},
      eprint={2310.06825},
      archivePrefix={arXiv},
      primaryClass={cs.CL},
      url={https://arxiv.org/abs/2310.06825}, 
}

@inproceedings{
xiao2024efficient,
title={Efficient Streaming Language Models with Attention Sinks},
author={Guangxuan Xiao and Yuandong Tian and Beidi Chen and Song Han and Mike Lewis},
booktitle={The Twelfth International Conference on Learning Representations},
year={2024},
url={https://openreview.net/forum?id=NG7sS51zVF}
}

@inproceedings{
fu2025not,
title={Not All Heads Matter: A Head-Level {KV} Cache Compression Method with Integrated Retrieval and Reasoning},
author={Yu Fu and Zefan Cai and Abedelkadir Asi and Wayne Xiong and Yue Dong and Wen Xiao},
booktitle={The Thirteenth International Conference on Learning Representations},
year={2025},
url={https://openreview.net/forum?id=FJFVmeXusW}
}

@inproceedings{
xiao2025duoattention,
title={DuoAttention: Efficient Long-Context {LLM} Inference with Retrieval and Streaming Heads},
author={Guangxuan Xiao and Jiaming Tang and Jingwei Zuo and junxian guo and Shang Yang and Haotian Tang and Yao Fu and Song Han},
booktitle={The Thirteenth International Conference on Learning Representations},
year={2025},
url={https://openreview.net/forum?id=cFu7ze7xUm}
}

@inproceedings{
tang2025razorattention,
title={RazorAttention: Efficient {KV} Cache Compression Through Retrieval Heads},
author={Hanlin Tang and Yang Lin and Jing Lin and Qingsen Han and Danning Ke and Shikuan Hong and Yiwu Yao and Gongyi Wang},
booktitle={The Thirteenth International Conference on Learning Representations},
year={2025},
url={https://openreview.net/forum?id=tkiZQlL04w}
}

@article{agrawal2023sarathi,
  title={Sarathi: Efficient llm inference by piggybacking decodes with chunked prefills},
  author={Agrawal, Amey and Panwar, Ashish and Mohan, Jayashree and Kwatra, Nipun and Gulavani, Bhargav S and Ramjee, Ramachandran},
  journal={arXiv preprint arXiv:2308.16369},
  year={2023}
}

@inproceedings{patel2024splitwise,
  title={Splitwise: Efficient generative llm inference using phase splitting},
  author={Patel, Pratyush and Choukse, Esha and Zhang, Chaojie and Shah, Aashaka and Goiri, {\'I}{\~n}igo and Maleki, Saeed and Bianchini, Ricardo},
  booktitle={2024 ACM/IEEE 51st Annual International Symposium on Computer Architecture (ISCA)},
  pages={118--132},
  year={2024},
  organization={IEEE}
}

@inproceedings{yu2022orca,
  title={Orca: A distributed serving system for $\{$Transformer-Based$\}$ generative models},
  author={Yu, Gyeong-In and Jeong, Joo Seong and Kim, Geon-Woo and Kim, Soojeong and Chun, Byung-Gon},
  booktitle={16th USENIX Symposium on Operating Systems Design and Implementation (OSDI 22)},
  pages={521--538},
  year={2022}
}

@inproceedings{lee2024infinigen,
  title={$\{$InfiniGen$\}$: Efficient generative inference of large language models with dynamic $\{$KV$\}$ cache management},
  author={Lee, Wonbeom and Lee, Jungi and Seo, Junghwan and Sim, Jaewoong},
  booktitle={18th USENIX Symposium on Operating Systems Design and Implementation (OSDI 24)},
  pages={155--172},
  year={2024}
}

@inproceedings{
gorle2025quantifying,
title={Quantifying Information Gain and Redundancy in Multi-Turn {LLM} Conversations},
author={Abhiram Rao Gorle and Amit Kumar Singh Yadav and Tsachy Weissman},
booktitle={First Workshop on Multi-Turn Interactions in Large Language Models},
year={2025},
url={https://openreview.net/forum?id=5gpABTkcUJ}
}

@misc{li2025singleturnsurveymultiturninteractions,
      title={Beyond Single-Turn: A Survey on Multi-Turn Interactions with Large Language Models}, 
      author={Yubo Li and Xiaobin Shen and Xinyu Yao and Xueying Ding and Yidi Miao and Ramayya Krishnan and Rema Padman},
      year={2025},
      eprint={2504.04717},
      archivePrefix={arXiv},
      primaryClass={cs.CL},
      url={https://arxiv.org/abs/2504.04717}, 
}

@inproceedings{zhong2024memorybank,
  title={Memorybank: Enhancing large language models with long-term memory},
  author={Zhong, Wanjun and Guo, Lianghong and Gao, Qiqi and Ye, He and Wang, Yanlin},
  booktitle={Proceedings of the AAAI Conference on Artificial Intelligence},
  volume={38},
  pages={19724--19731},
  year={2024}
}

@article{chhikara2025mem0,
  title={Mem0: Building production-ready ai agents with scalable long-term memory},
  author={Chhikara, Prateek and Khant, Dev and Aryan, Saket and Singh, Taranjeet and Yadav, Deshraj},
  journal={arXiv preprint arXiv:2504.19413},
  year={2025}
}

@misc{kim2025epicacheepisodickvcache,
      title={EpiCache: Episodic KV Cache Management for Long Conversational Question Answering}, 
      author={Minsoo Kim and Arnav Kundu and Han-Byul Kim and Richa Dixit and Minsik Cho},
      year={2025},
      eprint={2509.17396},
      archivePrefix={arXiv},
      primaryClass={cs.CL},
      url={https://arxiv.org/abs/2509.17396}, 
}

@inproceedings{
hu2026evaluating,
title={Evaluating Memory in {LLM} Agents via Incremental Multi-Turn Interactions},
author={Yuanzhe Hu and Yu Wang and Julian McAuley},
booktitle={The Fourteenth International Conference on Learning Representations},
year={2026},
url={https://openreview.net/forum?id=DT7JyQC3MR}
}

@inproceedings{
ghadia2025dialogue,
title={Dialogue Without Limits: Constant-Sized {KV} Caches for Extended Response in {LLM}s},
author={Ravi Ghadia and Avinash Kumar and Gaurav Jain and Prashant J. Nair and Poulami Das},
booktitle={Forty-second International Conference on Machine Learning},
year={2025},
url={https://openreview.net/forum?id=SuYO70ZxZX}
}

@inproceedings{zhang2025jenga,
  title={JENGA: Effective memory management for serving LLM with heterogeneity},
  author={Zhang, Chen and Du, Kuntai and Liu, Shu and Kwon, Woosuk and Mo, Xiangxi and Wang, Yufeng and Liu, Xiaoxuan and You, Kaichao and Li, Zhuohan and Long, Mingsheng and others},
  booktitle={Proceedings of the ACM SIGOPS 31st Symposium on Operating Systems Principles},
  pages={446--461},
  year={2025}
}

@article{dong2024hymba,
  title={Hymba: A hybrid-head architecture for small language models},
  author={Dong, Xin and Fu, Yonggan and Diao, Shizhe and Byeon, Wonmin and Chen, Zijia and Mahabaleshwarkar, Ameya Sunil and Liu, Shih-Yang and Van Keirsbilck, Matthijs and Chen, Min-Hung and Suhara, Yoshi and others},
  journal={arXiv preprint arXiv:2411.13676},
  year={2024}
}

@article{lieber2024jamba,
  title={Jamba: A hybrid transformer-mamba language model},
  author={Lieber, Opher and Lenz, Barak and Bata, Hofit and Cohen, Gal and Osin, Jhonathan and Dalmedigos, Itay and Safahi, Erez and Meirom, Shaked and Belinkov, Yonatan and Shalev-Shwartz, Shai and others},
  journal={arXiv preprint arXiv:2403.19887},
  year={2024}
}

@article{team2024gemma,
  title={Gemma 2: Improving open language models at a practical size},
  author={Team, Gemma and Riviere, Morgane and Pathak, Shreya and Sessa, Pier Giuseppe and Hardin, Cassidy and Bhupatiraju, Surya and Hussenot, L{\'e}onard and Mesnard, Thomas and Shahriari, Bobak and Ram{\'e}, Alexandre and others},
  journal={arXiv preprint arXiv:2408.00118},
  year={2024}
}

@inproceedings{oren-etal-2024-transformers,
    title = "Transformers are Multi-State {RNN}s",
    author = "Oren, Matanel  and
      Hassid, Michael  and
      Yarden, Nir  and
      Adi, Yossi  and
      Schwartz, Roy",
    editor = "Al-Onaizan, Yaser  and
      Bansal, Mohit  and
      Chen, Yun-Nung",
    booktitle = "Proceedings of the 2024 Conference on Empirical Methods in Natural Language Processing",
    month = nov,
    year = "2024",
    address = "Miami, Florida, USA",
    publisher = "Association for Computational Linguistics",
    url = "https://aclanthology.org/2024.emnlp-main.1043/",
    doi = "10.18653/v1/2024.emnlp-main.1043",
    pages = "18724--18741"
}

@article{devoto2025expected,
  title={Expected attention: Kv cache compression by estimating attention from future queries distribution},
  author={Devoto, Alessio and Jeblick, Maximilian and J{\'e}gou, Simon},
  journal={arXiv preprint arXiv:2510.00636},
  year={2025}
}

@article{park2026keydiff,
  title={Keydiff: Key similarity-based kv cache eviction for long-context llm inference in resource-constrained environments},
  author={Park, Junyoung and Jones, Dalton and Morse, Matthew and Goel, Raghavv and Lee, Mingu and Lott, Christopher},
  journal={Advances in Neural Information Processing Systems},
  volume={38},
  pages={5983--6019},
  year={2026}
}

@inproceedings{NEURIPS2019_2c601ad9,
 author = {Michel, Paul and Levy, Omer and Neubig, Graham},
 booktitle = {Advances in Neural Information Processing Systems},
 editor = {H. Wallach and H. Larochelle and A. Beygelzimer and F. d\textquotesingle Alch\'{e}-Buc and E. Fox and R. Garnett},
 pages = {},
 publisher = {Curran Associates, Inc.},
 title = {Are Sixteen Heads Really Better than One?},
 url = {https://proceedings.neurips.cc/paper_files/paper/2019/file/2c601ad9d2ff9bc8b282670cdd54f69f-Paper.pdf},
 volume = {32},
 year = {2019}
}

@inproceedings{voita-etal-2019-analyzing,
    title = "Analyzing Multi-Head Self-Attention: Specialized Heads Do the Heavy Lifting, the Rest Can Be Pruned",
    author = "Voita, Elena  and
      Talbot, David  and
      Moiseev, Fedor  and
      Sennrich, Rico  and
      Titov, Ivan",
    editor = "Korhonen, Anna  and
      Traum, David  and
      M{\`a}rquez, Llu{\'i}s",
    booktitle = "Proceedings of the 57th Annual Meeting of the Association for Computational Linguistics",
    month = jul,
    year = "2019",
    address = "Florence, Italy",
    publisher = "Association for Computational Linguistics",
    url = "https://aclanthology.org/P19-1580/",
    doi = "10.18653/v1/P19-1580",
    pages = "5797--5808",
}

@inproceedings{
wu2025retrieval,
title={Retrieval Head Mechanistically Explains Long-Context Factuality},
author={Wenhao Wu and Yizhong Wang and Guangxuan Xiao and Hao Peng and Yao Fu},
booktitle={The Thirteenth International Conference on Learning Representations},
year={2025},
url={https://openreview.net/forum?id=EytBpUGB1Z}
}

\appendix

\twocolumn[{%
    \centering
    \includegraphics[width=\textwidth]{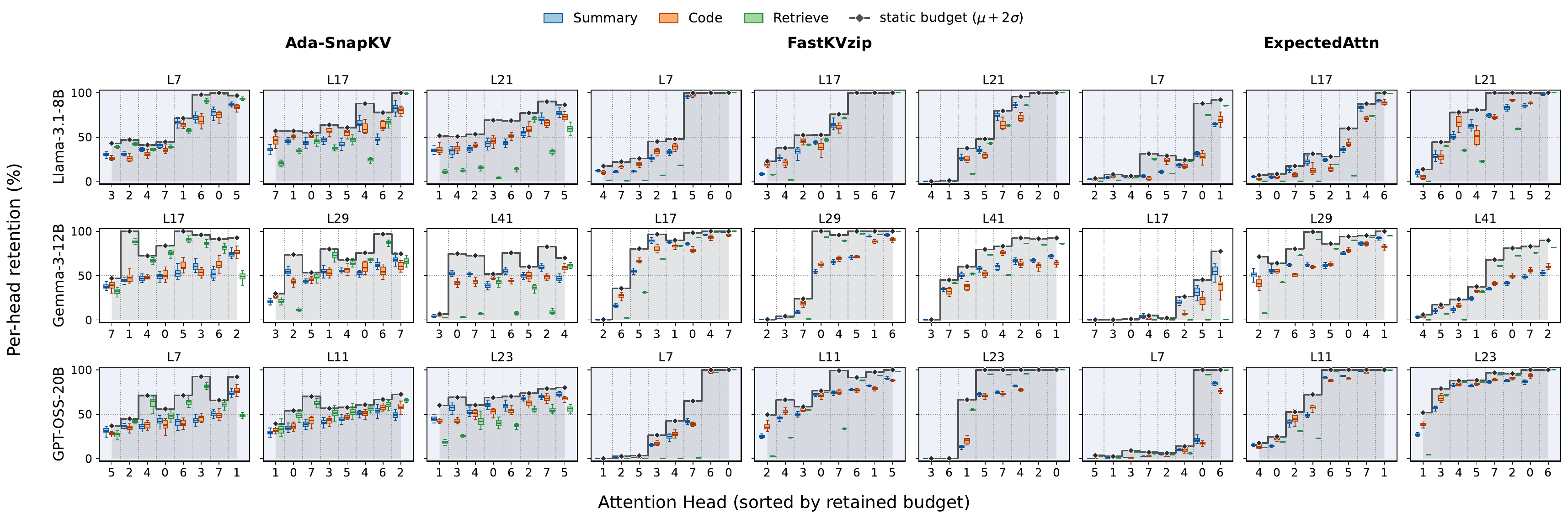}
    \captionof{figure}{Per-head KV retention rates (\%) under three non-uniform compression methods---Ada-SnapKV~\cite{li2024snapkv,feng2024ada}, FastKVzip~\cite{kim2026fast}, and Expected Attention~\cite{devoto2025expected}---at a 50\% target global budget, across three model families (rows: Llama-3.1-8B, Gemma-3-12B, GPT-OSS-20B) and three SCBench~\cite{li2024scbench} tasks including summarization (Summary), code understanding (Code; RepoQA), and fact retrieval (Retrieve; QA-Eng), shown for three layers per model; for the hybrid models (Gemma-3-12B, GPT-OSS-20B), these are full-attention layers. Within each panel, heads are sorted by retained budget, and each box shows the spread of one head's retention across the input samples of one task.}
    \label{fig:methods_3x9}
    \vspace{\baselineskip}
}]

\section{Retention Regularity across Compression Methods}
\label{sec:appendix_methods}

Figure~\ref{fig:kvsize} (\S\ref{key_observation}) establishes the two-level structural regularity of head-wise KV retention using KVzip~\cite{kim2025kvzip} as the scoring method. Figure~\ref{fig:methods_3x9} repeats the same rank-box analysis for the three non-uniform compression methods integrated in our evaluation (\S\ref{sec:eval-setup})---Ada-SnapKV, FastKVzip, and Expected Attention---across the same three model families and SCBench tasks.

The regularity is consistently reproduced for every method: within each panel, each head's per-task boxes are narrow and remain aligned across tasks, confirming that the head ranking is input-invariant and that each head's absolute retention ratio varies only within a narrow band. Accordingly, the overlaid static budget ($\mu + 2\sigma$) covers each head's observed spread while spending only a marginal amount of extra budget, for every method. At the same time, the \emph{shape} of the retention profile is clearly method-specific: Ada-SnapKV spreads the budget relatively evenly across heads, whereas FastKVzip and Expected Attention concentrate retention on a small subset of heads, leaving the remaining heads heavily pruned. This is precisely the setting \textsc{Tangram} is designed for: budget calibration (\S\ref{sec:calibration}) is performed per model and per method, so it faithfully captures whatever retention profile a scoring method induces, while the input-invariance demonstrated here guarantees that the offline-calibrated budgets transfer to unseen inputs. The key observation in \S\ref{key_observation} is therefore a property of non-uniform KV compression itself, not an artifact of a particular scoring method.

\end{document}